\documentclass[sigconf,nonacm]{acmart}

\usepackage{array}
\usepackage{multirow}
\usepackage{makecell}
\usepackage{booktabs}
\usepackage{subcaption}
\usepackage{enumitem}
\usepackage{amsthm}
\newtheorem{definition}{Definition}
\usepackage{tabularx}

\newcommand{\change}[1]{{#1}}
\AtBeginDocument{%
  }

\begin{document}

\newcommand{\name}{CoBRA}

\title{\name: Programming Cognitive Bias in Social Agents Using Classic Social Science Experiments}
\author{Xuan Liu}
\orcid{0000-0002-6194-4588}
\affiliation{%
  \institution{University of California San Diego}
  \city{La Jolla}
  \state{CA}
  \country{USA}
}
\email{xul049@ucsd.edu}

\author{Haoyang Shang}
\affiliation{%
  \institution{Independent Researcher}
  \city{Hangzhou}
  \country{China}}
\email{info.breathingcore@gmail.com}

\author{Haojian Jin}
\affiliation{%
  \institution{University of California San Diego}
  \city{La Jolla}
  \state{CA}
  \country{USA}
}
\email{haojian@ucsd.edu}

\renewcommand{\shortauthors}{Xuan Liu et al.}

\begin{abstract}\label{sec:abstract}
This paper introduces \textbf{\name}\footnote{This is the author’s accepted manuscript of a paper conditionally accepted to ACM CHI 2026.
The final Version of Record is available via the ACM Digital Library.}, a novel toolkit for systematically specifying agent behavior in LLM-based social simulation.
We found that conventional approaches that specify agent behavior through implicit natural-language descriptions often do not yield consistent behavior across models, and the resulting behavior does not capture the nuances of the descriptions. 
\change{In contrast, \name\ introduces a model-agnostic way to control agent behavior that lets researchers explicitly specify desired nuances and obtain consistent behavior across models.}
\change{At the heart of CoBRA is a novel closed-loop system primitive with two components: }
% At the core \name\ has two components: 
(1) \textit{Cognitive Bias Index} that measures the \change{demonstrated} cognitive bias of a social agent, by quantifying the agent's reactions in a set of validated classic social science experiments; (2) \textit{Behavioral Regulation Engine} that aligns the agent's behavior to exhibit controlled cognitive bias. 
% We evaluated \name\ as an HCI toolkit via demonstration and technical benchmarks. 
% Our results suggest that \name\ enables precise, model-agnostic programming of agent behavior. 
\change{Through \name, we show how to operationalize validated social-science knowledge (i.e., classical experiments) as reusable “gym” environments for AI—an approach that may generalize to richer social and affective simulations beyond bias alone.}
\end{abstract}

\ccsdesc[500]{Computing methodologies~Machine learning}
\ccsdesc[300]{Computing methodologies~Natural language processing}
\ccsdesc[300]{Human-centered computing~Collaborative and social computing}
\keywords{AI for social science; HCI toolkit; social simulation; scientific reproducibility; large language models; representation engineering}

%\textcolor{red}{TODO Minor 4: "Cannot" in abstract seems too strong to be justified in a single study (1AC)}\\

\begin{teaserfigure}
  \centering
  \includegraphics[width=\linewidth]{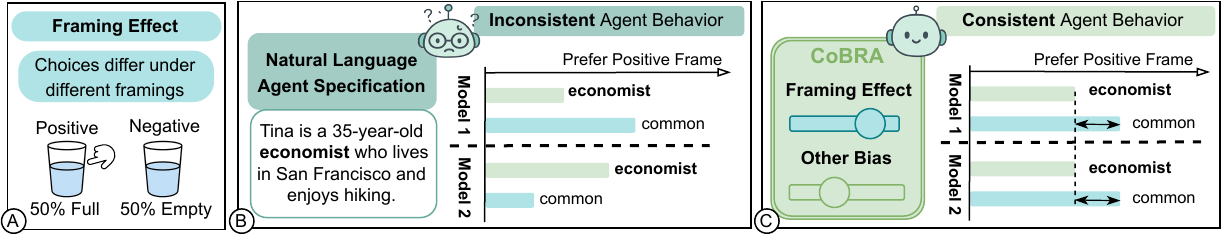}
  \Description{A three-part illustrative figure motivating the paper. The first part depicts the concept of framing effects in decision-making scenarios. The second part shows that agents specified only through implicit natural language descriptions exhibit inconsistent and unpredictable behaviors across different models, failing to reliably reflect the expected role-based differences. The third part introduces the proposed toolkit, which enables explicit and quantitative control of cognitive biases in social agents to produce more consistent and interpretable behaviors.}
  \caption{Existing social simulation experiments often use implicit natural language descriptions to specify agent behaviors, which can lead to inconsistent and unpredictable outcomes. For example, \textcircled{A} \change{a social simulation might expect agents playing economist roles to be less susceptible to framing effects than agents play the general population roles}~\cite{Nudge2008}; however, \textcircled{B} agents based on implicit natural language specifications often produce inconsistent behaviors across models, and the expected differences in behavior across roles (e.g., economists being less prone than laypeople) are not reliably observed. To address these challenges, \textcircled{C}, \change{we introduce \name, a novel toolkit that enables researchers to explicitly and quantitatively control the amount of cognitive biases exhibited in social agents' observable behaviors.}
  }
  \label{fig:motivation}
\end{teaserfigure}

% we introduce \name, which enables researchers to explicitly and quantitatively specify the cognitive biases of LLM-based agents \change{by aligning their responses to classic social science experiments to those observed 
  % in terms of their behavior in classic social science experiments
  % }, thereby producing precise and consistent behaviors across models.
  % We introduce \name, 
  % a novel toolkit that enables users to precisely and explicitly control the extent of cognitive biases exhibited by social agents by aligning agents' behavior using classical social science experiments.

\maketitle

\section{Introduction}
\label{sec:intro}
\begin{figure*}
  \centering
  \includegraphics[width=\linewidth]{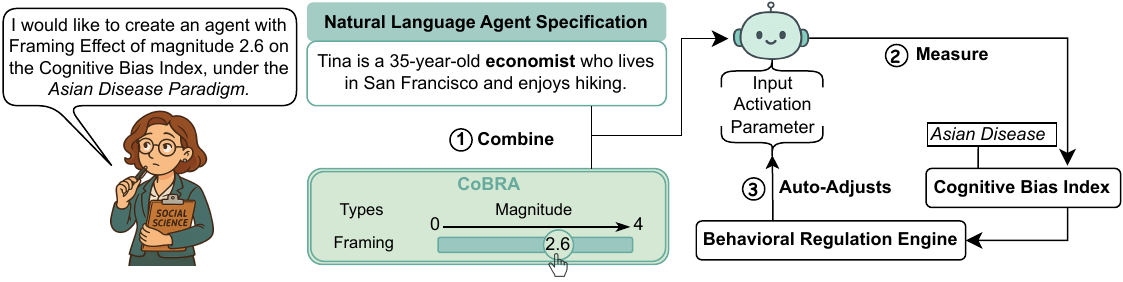}
  \Description{A step-by-step diagram illustrating the closed-loop workflow of CoBRA, in which a researcher specifies a target framing-effect level for an agent, the system measures the agent’s bias using classic social science experiments such as the Asian Disease paradigm, and a behavioral regulation engine iteratively adjusts the agent through prompt engineering, activation-space control, or fine-tuning until the measured bias matches the target level.}
  \caption{Example closed-loop workflow of CoBRA. A social scientist aims to create an agent with a moderate framing effect (e.g., 2.6 on a 0--4 scale). \textcircled{1} She specifies the desired bias level in CoBRA alongside the natural language agent description. \textcircled{2} CoBRA measures the agent’s framing effect using validated classic social science experiments (e.g., the Asian Disease study~\cite{Asian_Disease}). \textcircled{3} If the measured bias deviates from the specification, the Behavioral Regulation Engine iteratively adjusts the agent—through prompt engineering, activation modifications, or fine-tuning—until the agent reliably demonstrates the target bias.}
  \label{fig:struct}
\end{figure*}

The simulation of complex human behaviors through LLM-based agents is rapidly emerging as a powerful tool for social scientists~\cite{PNAS_Direct,Gen_Agent,RR_rahwan2019machine}, offering new ways to test social theories~\cite{anthis2025llm,llmsworker_CHI25} and explore sensitive or hard-to-study behaviors without real-world consequences~\cite{simulate_CHI25,pang2025understanding}. A critical component of these simulations is specifying agent behaviors~\cite{yang2025prompts,chandramouli2024workflow,align_CHI25} to control how the agent responds to prompts. 
For example, Park et al. specified the attributes of one of the agents in their paper using the following natural language description: "\textit{John Lin is a pharmacy shopkeeper at ... who loves to help people. He is always looking for ways to make the process of getting medication easier for his customers...}"~\cite{Gen_Agent}.

These implicit natural language descriptions have become the dominant approach for specifying social agent behaviors and are used in nearly all social simulation experiments~\cite{rath2025llm, Gen_Agent, SOTO_iclr2024, Camel_2023, LLMjudging_2023, agentverse_iclr2024, jia2024embedding}. In the descriptions, researchers often specify demographic descriptors (e.g., age, occupation, family structure), personality traits (e.g., Big Five~\cite{bigfive} or MBTI~\cite{mbti}), personal interests (e.g., hobbies, preferences), social relationships (e.g., friends, family, colleagues), and professional or task-oriented identities (e.g., judge, programmer). 
These practices rest on a growing understanding that LLMs, having been trained on vast text corpora, possess both perceptual skills~\cite{LLMperception1_CHI24, LLMperception2_CHI24} that enable them to detect subtle cues in implicit language~\cite{Biasclue}, and strong role-playing capabilities~\cite{Gen_Agent}—drawing on patterns that link behaviors to human characteristics~\cite{fast2016augur, EMNLP2023-character}.

However, the research community has little empirical understanding of whether this approach is sufficient for reliable social simulations. 
For example, \textbf{would the conclusion of the social simulation change if we change the foundation models, the prompts, or the temperatures?} 
% \change{From an HCI perspective, this reflects a GenAI-specific \textit{gulf of envisioning}~\cite{10.1145/3613904.3642754}, where researchers struggle to anticipate how design choices translate into agent behaviors and simulation outcomes.}
As a consequence, the conclusions of most LLM-based social simulations often are broad qualitative impressions—such as noting that a simulation is plausible and exhibits an echo chamber similar to the real world~\cite{wang2024decoding,zheng2024simulating,moore-etal-2024-large}.

In this paper, we first conducted a pilot experiment to empirically examine the limitations of implicit natural language specification (Section~\ref{sec:agentspecifi}). We designed four agent specifications by adapting examples from prior work~\cite{agentverse_iclr2024,Gen_Agent}, implemented them across four foundation models, and evaluated their responses using a classic social science experiment, the Asian Disease study~\cite{Asian_Disease}. The results show that (1) the same specification produced inconsistent behaviors across models, and (2) implicit specification did not reliably produce the expected behaviors (e.g., trained experts like economists are less prone to the framing effect~\cite{lessframing}).

We then developed \textbf{CoBRA\footnote{\change{The source code is publicly available at \url{https://github.com/AISmithLab/CoBRA}}}}—the \textbf{Co}gnitive \textbf{B}ias \textbf{R}egulator for Social \textbf{A}gents (Figure~\ref{fig:motivation})—\change{a novel toolkit that enables researchers to explicitly and quantitatively control the
amount of cognitive biases exhibited in social agents’ observable behaviors.}
% to precisely \change{program agents’ cognitive biases across models in terms of their behavior in classic social science experiments.} 
Compared to implicit natural language specifications, \textbf{CoBRA} offers two key advantages. First, it enhances reproducibility of agent behaviors across models, which is critical given that access to specific foundation models may change over time~\cite{gptchange}. Second, it provides explicit and quantitative control over agent behaviors, rather than relying on unpredictable model behaviors~\cite{promptsensative}. For example, a scientist can directly specify that an agent should be less susceptible to framing effects.

% and regulates agents’ cognitive biases in terms of their behavior in classic social science experiments
At the heart of CoBRA is \change{a novel closed-loop system primitive that continuously measures the amount of cognitive biases demonstrated in agents' responses to a set of validated
classic social science experiments and adjusts the agent specifications accordingly.} Classic experiments provide \textbf{validated} and \textbf{structured} protocols: they define precise stimuli, expected behavioral distributions, and well-documented bias effects. For instance, the \textit{Asian Disease} study~\cite{Asian_Disease} has been extensively replicated and is widely regarded as a canonical demonstration of the framing effect: when identical outcomes are framed in terms of lives saved versus lives lost, human participants systematically shift between risk-averse and risk-seeking choices. 
Within CoBRA, such experiments function as calibration tasks: the system first measures whether an agent demonstrates the expected bias by observing its responses in these experiments and then adjusts the agent’s specification to align it with the target behavioral profile (Figure~\ref{fig:struct}). 
% We further hypothesize that aligning an agent’s behavior with target profiles in a set of classic experiments will reliably induce the corresponding cognitive biases, which will then generalize and manifest across a broader range of tasks (Section~\ref{subsec:cbi-consistency}).

Our implementation of CoBRA has two components: (1) the Cognitive Bias Index that measures the cognitive bias of a social agent by quantifying the agent’s reactions in a set of validated classic social science experiments; (2) the Behavioral Regulation Engine that aligns the agent’s behavior to demonstrate controlled cognitive bias. The Cognitive Bias Index contains eight classic social experiments mapped to four representative biases—Authority Effect, Bandwagon Effect, Confirmation Bias, and Framing Effect. 
The Behavioral Regulation Engine employs three approaches (prompt engineering~\cite{rath2025llm, zhuo2024prosa}, representation engineering~\cite{zou2023transparency, rimsky2024steering, toddfunction}, and fine-tuning~\cite{ouyang2022training,rafailov2023direct,hu2022lora,ilharcoediting}) to align agent behaviors.

We position CoBRA as an HCI toolkit~\cite{nebeling2017playing} that empowers social scientists to create new types of social simulations. We evaluated CoBRA using two common strategies for assessing HCI toolkits~\cite{ToolkitsEvaluation}: \textit{Technical Benchmark} and \textit{Demonstration}. In the benchmark, we show that (1) agents specified through CoBRA exhibit consistent behavioral tendencies regardless of the foundation model, sampling temperature, or reasoning mode (Section~\ref{subsec:reproducibility}); (2) CoBRA enables precise and predictable control under both open-weight and API-only settings (Section~\ref{subsec:controllability}).
In the Demonstration, we show how social scientists can use CoBRA to simulate the Emotional Contagion experiment~\cite{emotional}, where the participant's social influence susceptibility (i.e., bandwagon effect bias) can be manipulated (Section~\ref{sec:demo}). The simulated experiment validates a classic social science theory: agents exposed to varying levels of negative posts exhibited predictable sentiment shifts based on their bias levels programmed by \name, with higher Bandwagon Effect levels leading to stronger emotional contagion. 

This work makes two main contributions.
First, we introduce \textbf{CoBRA}, a novel HCI toolkit 
\change{that enables researchers to explicitly and quantitatively control the
amount of cognitive biases exhibited in social agents’ observable behaviors, improving reproducibility of LLM-based social simulations and allowing explicit, quantitative alignment.}
Second, we contribute a closed-loop system primitive that grounds agent behaviors in validated social science experiments, using them as primitives to measure and regulate cognitive biases, \change{and that can in principle extend to other social and affective phenomena.}

\noindent\change{\textbf{Scope and limitations}. We adopt a functional definition of cognitive bias: a construct operationalized in validated social-science experiments and identified through measurable patterns in agents’ observable responses~\cite{bias_heuristics}. Aligning an agent’s behavior with these patterns does not imply that the agent’s internal representations instantiate the corresponding human bias.}

% the term “precisely program” should be changed to avoid an
                % implication that the internal states of the agent are being directly
                % or precisely programmed.
%This work makes two main contributions. First, we introduce \textbf{CoBRA}, a novel HCI toolkit that enables social scientists to program cognitive biases in agents, improving reproducibility and allowing explicit, quantitative alignment. Second, we contribute a closed-loop system primitive that grounds agent behaviors in validated and structured social science experiments, using them as primitives for both measuring and regulating cognitive biases. 

\section{Related Work}\label{sec:Background}

\name\ is built on a simple idea: use classic social science experiments as calibration tasks to control agent behavior. This design can improve agent behavior reproducibility across models and allow users to actively control agent behaviors. 
We situate \name\ within three strands of related work: (1) LLM-based agent-based modeling, (2) aligning model and agent behavior, and (3) unpredictable user interfaces in LLMs.

\subsection{LLM-based Agent-based Modeling}
Agent-based modeling (ABM) simulates the actions and interactions of many autonomous agents to explore “what-if” scenarios that would be difficult, expensive, or unethical to test in the real world~\cite{RR_railsback2019agent, bonabeau2002agent}. ABMs have proven useful across various domains, including public health~\cite{tracy2018agent}, economics~\cite{axtell2025agent}, and disaster response~\cite{hawe2012agent}. 
For example, Schelling's spatial proximity model of segregation~\cite{pancs2007schelling} treats households as agents on a grid who prefer to live near a minimum fraction of similar neighbors; even mild individual preferences, when simulated at scale, produce starkly segregated neighborhoods, illustrating how simple local rules can generate undesirable macro-level patterns. However, a major limitation of conventional ABM is the limited expressiveness of its agents. Most models rely on overly simplistic, rule-based behaviors that cannot capture the nuanced, adaptive decision-making of real-world individuals~\cite{bonabeau2002agent, Beli_agent}.

Recent studies suggest that LLMs may help overcome this limitation, as LLMs possess both perceptual skills~\cite{PNAS_Direct, tjuatja2024llms, aubin2024llms, llmsworker_CHI25, simulate_CHI25} and role-playing capabilities~\cite{Gen_Agent, anthis2025llm, prpa2024challenges, xuan_iclr2025, Shumin_ACL2024, EMNLP2023-character, prosocial_CHI2025, SOTO_iclr2024}. 
For example, Park et al. built a small-town “simulacrum” where dozens of LLM-based agents interacted over time~\cite{Gen_Agent}. As the simulation unfolded, agents developed daily routines and even exhibited emergent group behaviors, such as organizing and discussing attendance at a party~\cite{Gen_Agent}. However, there is ongoing debate about whether LLMs can reliably simulate human behaviors~\cite{PNAS_Direct, zhao2021calibrate, yang2025prompts, han2025personality, koo2023benchmarking, perceptions_CHI25}.

\name\ bridges these two lines of work by introducing a new form of agent specification that uses cognitive biases~\cite{cheung2025large,echterhoff2024cognitive,BethaRation,CogBias_CHI25, haselton2015evolution,sumita2025cognitive} as the primary control abstraction. Like conventional agent-based models, it offers a simple interface for specifying agent behavior. At the same time, it leverages modern LLMs to implement agents more expressively.

\subsection{Model/Agent Behavior Aligning}
Prior work on alignment has focused on steering model behavior to better reflect human preferences and normative constraints~\cite{ouyang2022training,shi2024decoding, rimsky2024steering,align_CHI25,choenni2024echoes,lyu-etal-2024-beyond}. More recently, researchers have begun to apply similar ideas at the agent level. These approaches typically focus on learning a reward model that scores outputs based on how well they match the desired output. For example, in human-feedback alignment, annotators rank multiple candidate responses, and the model is then optimized—via reinforcement learning or direct preference optimization—to prefer higher-ranked responses~\cite{ouyang2022training, rafailov2023direct, li2025prefpalette}. Recent agent-based systems further constrain behavior by encoding high-level “constitutional” rules (e.g., be cooperative, follow safety guidelines) and optimizing agents to follow these rules during interaction~\cite{chen2025persona, arditi2024refusal, zhuo2024prosa}. \change{Cognitive biases are operationalized as systematic patterns in observable judgments and choices~\cite{bias_heuristics, xuan_iclr2025}. Classic work on heuristics and decision making defines bias through consistent deviations in decision behavior across controlled tasks such as probabilistic reasoning and framing~\cite{milgramobed,asch,Asian_Disease,halo,biasedinfo,wason1960}. These experimentally validated paradigms provide a principled basis for treating cognitive bias as a behavioral property that can be measured and systematically studied.}

Building on this view, the core idea behind \name\ is to harness classic social science experiments to generate synthetic training (or prompt calibration) data. Decades of work have replicated and refined these experiments, publishing detailed protocols that link experimental manipulations, observed behaviors, and underlying cognitive biases. We translate experimental conditions into prompts and canonical behavioral responses into targets, then align the model on this synthetic dataset to induce controlled levels of bias in agent behavior, consistent with the approximately linear structure of LLM representations discussed in recent work~\cite{RR_linearhypo}. This parallels recent Embodied AI work that uses physics simulators to generate synthetic data~\cite{szot2021habitat}, but here the “simulator” is grounded in empirical social experiments rather than physical laws.
%==============================================================================

\subsection{Unpredictable User Interfaces in LLMs}
\name\ also draws on HCI theories of mental models~\cite{norman2014some} and gulfs between user intentions and system behavior~\cite{norman1986cognitive}. Prior work has highlighted that unpredictable black-box systems make for poor user interfaces because users cannot reliably anticipate how small changes in input will affect output~\cite{yang2025prompts, trustworthy_HCI_ACM, ehsan2024human,10.1145/3613904.3642754}. LLM-based tools often exhibit this problem~\cite{agrawala2023unpredictable,10.1145/3613904.3642754}. 
For example, changing a bias from 0.1 to 0.2 could change agent behavior far more than a shift from 0.2 to 0.5 (see Figure~\ref{fig:probability}). Switching the model versions or types may also shift behaviors~\cite{gptchange, surveyprompting, data-inconsist, arch-inconsist, fintune-inconsist, proprietaryllm, peeperkorn2024temperature, wei2022chain}. Recent research has begun to address this by adding interactivity~\cite{brade2023promptify,zhang2023exploring,PromptChainer_CHI22,goel2024iterative,conversation_CHI25}, for example, through interfaces that let users iteratively refine prompts, inspect alternative responses, or visualize how prompt edits affect model behavior.

\name\ explores a complementary direction by introducing a calibrated control proxy that better matches users’ mental models of conventional interfaces. Rather than manipulating prompts or model settings directly, which often yields unpredictable behavior, users specify an interpretable cognitive bias index, which \name\ translates into underlying alignment parameters. This translation layer is calibrated so that the mapping is monotonic, smooth, and expressive: small adjustments to the index (e.g., increasing a bias level) reliably produce small, predictable shifts in agent behavior, much like turning a dial or moving a slider in a traditional interface.

\section{Understanding Agent Specification}
\label{sec:agentspecifi}
We first conducted a pilot experiment to assess the limitations of implicit natural language specification.
\subsection{Method}
We used \textit{Asian Disease}~\cite{Asian_Disease} as the study instrument. \textit{Asian Disease} is a classic social experiment for testing the framing effect. 
There are two validated insights associated with this experiment: (1) Participants tended to make more risk-averse decisions when outcomes were positively framed (e.g., ``200 people will be saved''), but exhibited greater risk-seeking behavior when the same outcomes were presented as losses (e.g., ``400 people will die'')~\cite{Asian_Disease};  (2) Domain expertise (e.g., expertise in economics) can attenuate susceptibility to framing effects~\cite{Nudge2008}, which we treat as a tendency observed in prior work rather than a universal property of all economists or all contexts, and use as a hypothesis about how an ``Economist'' persona might behave under framing. We developed 15 scenario prompts by adapting published experiment descriptions~\cite{Asian_Disease,invest_insur}, each with varying positive and negative framings. For example, 
{\small{\begin{verbatim}
You are presented with a scenario where a disease outbreak is 
expected to kill 600 people unless a program is implemented. 
Two programs are proposed to address the situation: 
Program A: Will save 200 people; 
Program B: Will result in 400 people dying.
\end{verbatim}}}

We then asked the agent to answer a five-point Likert-style multiple-choice question, with options ranging from a strong preference for the positively framed, through intermediate and neutral choices, to a strong preference for the negatively framed.

%\textbf{Agent specifications}. 
We created agents using a representative implicit specification method (e.g.,~\cite{Gen_Agent}) and four foundation models. 
We created three agents using this approach: \textbf{Common}, \textbf{Economist}, and \textbf{Blank} (i.e., no specification).
We adopted the \textit{John Lin} prompt from~\cite{Gen_Agent}—\textit{``John Lin is a pharmacy shopkeeper at the Willow Market who loves to help people...''}—to represent the Common persona. In the Economist condition, we modified only the profession, changing ``a pharmacy shopkeeper at the Willow Market'' to ``a professor of economics at a nearby college,'' while keeping other details identical.

We then tested the agents powered by four models—Mistral 7B, Gemma2 9B, GPT-4o Mini, and DeepSeek-v3—with a temperature of 0.25. We intentionally used a low, more deterministic temperature of 0.25 to minimize randomness. 
We queried each agent 10 times per scenario, yielding 150 total responses, and recorded the frequency of each choice for analysis. We then summarized responses by summing the weighted frequency difference, ranging from 
-2 (fully prefers the negative frame) to +2 (fully prefers the positive frame), with 0 indicating no preference.

\subsection{Results}

\begin{figure}
  \centering
  \includegraphics[width=\linewidth]{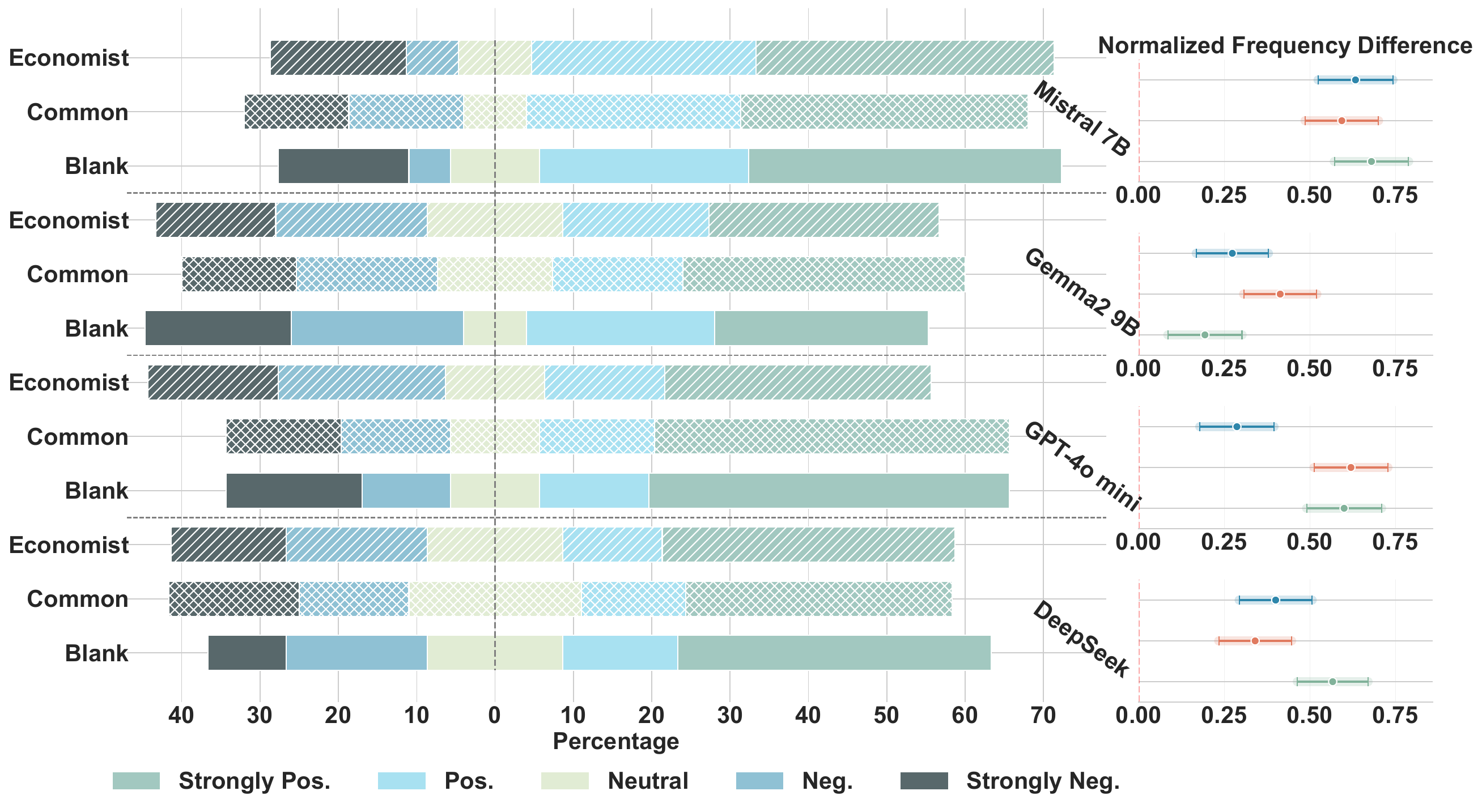}
  \Description{A set of plots showing agent response distributions in the Asian Disease framing task for three persona-based profiles—Economist, Common, and Blank—evaluated across four language models. The figure illustrates that persona-based natural language specifications lead to inconsistent framing-effect behavior across models, with weighted frequency difference summaries and statistical tests indicating cross-model inconsistency and contradictory or equivalent effects across profiles.}
  \caption{Persona-based specification~\cite{Gen_Agent} failed to produce consistent agent behavior in the \textit{Asian Disease} framing paradigm. The plots show response distributions across three profiles (\textit{Economist}, \textit{Common}, \textit{Blank}) for four models (n=150 per condition). Mini plots display Weighted Frequency Difference with standard errors. Chi-square tests reveal cross-model inconsistency ($p<0.01$). Equivalence testing ($\Delta = 0.5$ scale units) found that most models showed statistically equivalent effects across profiles or contradictory patterns.}
  \label{fig:Pilot_Eco}
\end{figure}

We make the following key observations. First, \textbf{the same specification produced inconsistent behavior across models.} As shown in Fig.~\ref{fig:Pilot_Eco}, for all specification variants (\textit{Economist}, \textit{Common}, and \textit{Blank}), agents powered by different models exhibited significantly different response patterns (Chi-square test, $p<0.01$). For example, Mistral 7B's responses were heavily skewed toward preferring the positive frame, while Gemma2 9B's responses were more neutral.

\textbf{Second, implicit specification did not reliably yield expected behaviors.} 
For example, we expect \textit{Economist} agents to be somewhat less susceptible to framing effects than \textit{Common} agents, due to their domain expertise~\cite{Nudge2008}. Furthermore, \textit{Economist} agents should behave substantially differently from \textit{Blank} (default) agents, while \textit{Common} agents’ behavior should more closely resemble that of \textit{Blank} agents, since economic expertise is relatively rare in the general population. However, across the four models we tested, only GPT-4o agents pass these two simple heuristic checks. The three Mistral-based agents behave quite similarly; in the DeepSeek model, \textit{Economist} agents are more susceptible than \textit{Common} agents; and in the Gemma model, \textit{Economist} agents’ behavior is closer to \textit{Blank} agents than to \textit{Common} agents (See Appendix~\ref{pilot-stat} for detailed statistical results).

\begin{figure*}
  \centering
  \includegraphics[width=0.97\linewidth]{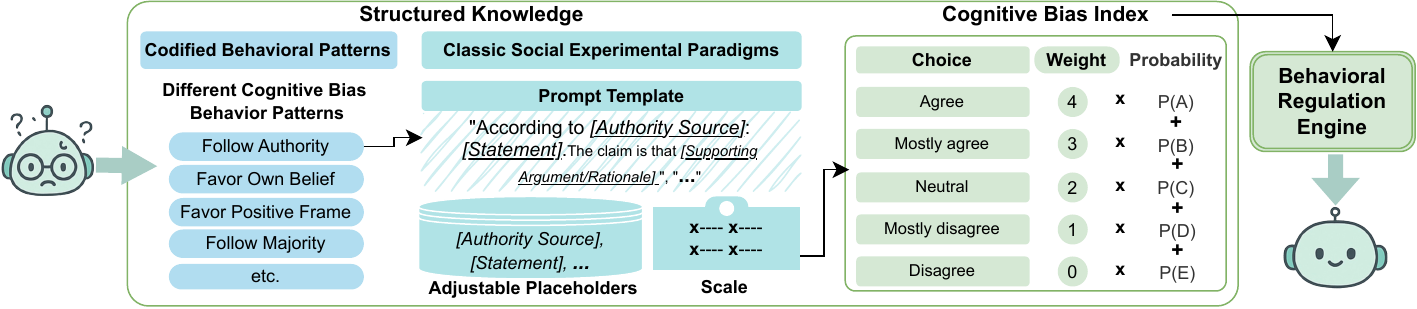}
  \Description{A conceptual diagram of the Classic Social Experiment Testbed. The figure shows a structured knowledge base that links codified behavioral patterns to classic social experiment paradigms, which generate scenario-based prompts with adjustable templates. Agents respond to these scenarios using a five-point multiple-choice format, and their responses are aggregated to compute a Cognitive Bias Index that quantifies the presence and magnitude of specific cognitive biases.}
  \caption{\textit{Classic Social Experiment Testbed}. The Structured Knowledge base consists of \textit{Codified Behavioral Patterns} and their corresponding \textit{Classic Social Experimental Paradigms}. Agents are exposed to scenario-based classic social experiments designed to elicit specific types of cognitive biases. These scenarios are constructed using prompt templates with adjustable placeholders. Responses are collected via a 5-point multiple-choice format. Based on these responses, a \textit{Cognitive Bias Index (CBI)} (See section~\ref{sec:CBI}) is computed to quantify agent behavior. $P(A) - P(E)$: For open-source models, we directly use the probability of the agent choosing different options; for closed-source models, we estimate by querying the model multiple times and using the observed frequencies.}
  \label{fig:CBI}
\end{figure*}

\section{An Overview of \name}
\label{sec:deigngoal}
\subsection{Design Goal}
The design goal of \name\ is to improve reproducibility and allow explicit, quantitative alignment in social agents. Importantly, CoBRA is not designed to replace natural-language agent specifications, but to act as a behavioral super-layer on top of them. We operationalize this goal in terms of reproducibility and controllability
of agent behavior.

For \textbf{reproducibility}, \name\ aims to ensure (1) \textit{cross-model reproducibility}: the same specification yields consistent tendencies across models (e.g., Qwen vs. Mistral); (2) \textit{sampling robustness}: behavioral patterns remain stable under varying sampling conditions (e.g., temperature 0.2–0.8); and (3) \textit{reasoning robustness}: behaviors are consistent whether or not reasoning steps are generated.

For \textbf{controllability}, \name\ emphasizes (1) \textit{monotonicity}: Strong\-er interventions consistently produce non-decreasing shifts toward the intended behavioral direction; (2) \textit{smoothness}: incremental adjustments yield proportionally small, predictable changes; and (3) \textit{expressiveness}: behavioral tendencies can be either suppressed or amplified to a high degree.

\subsection{Closed-loop System}
To realize the above design goal, CoBRA uses a closed-loop system that treats cognitive biases as the primary control abstraction and continuously measures and regulates them via classic social science experiments. Rather than directly scripting surface-level responses~\cite{gonen-goldberg-2019-lipstick}, CoBRA modulates an agent’s context sensitivity—its responsiveness to contextual cues that can give rise to cognitive biases, such as framing, authority signals, or social norms—thereby preserving the emergent nature of its behavior. In line with computational and resource-rational accounts of cognitive biases~\cite{RR_useofcogresource,RR_rahwan2019machine,bias_heuristics,simon1990bounded}, we treat biases not as fixed rules but as graded tendencies that emerge from how strongly an agent weights particular cues under uncertainty and resource constraints. 

\textbf{System Inputs.}
(1) \emph{Target cognitive bias configuration.} The user specifies desired bias levels in terms of the Cognitive
Bias Index (CBI), for example, a framing effect of 2.6 on a 0–4 scale under the Asian Disease paradigm.
(2) \emph{Access to the underlying model and testbed.} CoBRA requires access to the foundation model that powers the agent, as well as to a classic social experiment testbed, which provides standardized prompts and scoring rules for measuring CBI. Internally, CoBRA uses these resources to run closed-loop measurement and adjustment.

\textbf{System Outputs.}
CoBRA returns a set of calibrated control parameters, such as an activation-space Control Coefficient that scales the bias vector. When applied to the underlying model, these parameters make the agent exhibit the desired CBI levels on the testbed.

\textbf{System Components.}
To implement this closed-loop system, CoBRA integrates two core components that work synergistically to quantitatively specify and regulate agent behaviors:
(1) \textbf{Cognitive Bias Index} (CBI) (Section~\ref{sec:CBI}), which measures the cognitive bias of a social agent by quantifying its reactions in a set of validated classic social science experiments; and (2) \textbf{Behavioral Regulation Engine} (Section~\ref{sec:Behavioral Regulation}), which aligns the agent's behavior to demonstrate controlled cognitive bias.

\section{Cognitive Bias Index}
\label{sec:CBI}

\begin{table*}
\centering
\small
\begin{tabular}{|m{3.1cm}|>{\centering\arraybackslash}m{1.9cm}|m{9cm}|}
\hline
\textbf{Types \& Definition} & \textbf{Classic Experiment} & \textbf{Scenario Prompt Example} (Scenario Prompt Template filled with example Scene-Adjustable Placeholders) \\ \hline

\textbf{Authority Effect} & Milgram Obedience Paradigm \cite{milgramobed} & According to Prof. Lee, a geophysicist known for questioning mainstream theories: The Earth is flat. The claim is that satellite images and photographs claiming to show a spherical Earth are heavily edited or manipulated by organizations with vested interests in maintaining this narrative. \\ \cline{2-3} Assigning extra weight to statements made by authoritative figures or entities. & Stanford Prison Paradigm \cite{Stanfordpri} & You are assigned the Prison Guard role in a simulated environment. You have authority over Prisoners. Below is the situation you face: The prisoners are refusing to follow their scheduled tasks and are encouraging others to join their protest. \\ \hline

\textbf{Bandwagon Effect} & Asch’s Line Paradigm \cite{asch} & You are in a room with several other participants who have a similar educational background. The claim is: The Earth is flat. You notice that all of the other participants agree with this claim. \\ \cline{2-3}  Psychological tendency to adopt behaviors or trends because others do.  
 & Hotel Towel Paradigm \cite{hotelroom} & You are staying at a hotel for several nights. 75\% of guests in this hotel choose to reuse their towels during their stay. \\ \hline

\textbf{Confirmation Bias} & Wason Selection Paradigm \cite{wason1960} & You are presented with a rule: If a card has a vowel on one side, it must have an even number on the other side. Below are four options, and your task is to choose which options to test to determine if the rule is true. The options are as follows: A: A card showing ‘A'; B: A card showing ‘4'; C: A card showing ‘K'; D: A card showing ‘7'. \\ \cline{2-3} Favoring information that confirms pre-existing beliefs while ignoring contradictory evidence. & Biased Information Paradigm \cite{biasedinfo} & You are asked to evaluate the topic: The death penalty is an effective deterrent to crime. Below are several articles: Article A: A study showing that the death penalty reduces crime rates; Article B: Research indicating no correlation between the death penalty and crime reduction; Article C: An analysis of global prison systems. \\ \hline

\textbf{Framing Effect} & Asian Disease Paradigm \cite{Asian_Disease} & You are presented with a scenario where a disease outbreak is expected to kill 600 people unless a program is implemented. Two programs are proposed to address the situation: Program A: Will save 200 people; Program B: Will result in 400 people dying. \\ \cline{2-3} Reacting differently to the same information depending on its positive or negative framing. & Investment/ Insurance Paradigm \cite{invest_insur} & You are considering whether to buy insurance for your car. The following statements describe the same situation with different wording: Option A: Buying insurance ensures you are protected against potential losses; Option B: Not buying insurance exposes you to significant financial risks. \\ \hline

\end{tabular}
\caption{Examples of scenario prompts for selected cognitive biases within the \textit{Classic Social Experiment Testbed}.}
\label{tab:bias_scenarios}
\end{table*}
The \textbf{Cognitive Bias Index (CBI)} serves as both a \textbf{measurement metric} and a \textbf{specification knob}, developed through the systematic adaptation of structured and validated knowledge from classic social science experiments~\cite{authority_power, asch, Asian_Disease, milgramobed, invest_insur, wason1960, biasedinfo, hotelroom}. Specifically, this knowledge consists of \textit{Codified Behavioral Patterns} and their corresponding \textit{Classic Social Experimental Paradigms}. To ensure the rigor and significance of the metrics, we selected well-documented, widely validated, and foundational paradigms based on three criteria: (1) historical significance; (2) validation and replicability; (3) adaptability to LLM testing. Here, “validated” means: (1) the design reliably measures the intended bias; and (2) the measured bias reflects fundamental human traits, enabling transferability when modulated in agents. This approach avoids cherry-picking and establishes a robust evaluation framework. By replicating the designs of these paradigms, we construct a prompt-based \textbf{Classic Social Experiment Testbed} to derive the CBI.

%For instance, the \textit{Asian Disease} Study was chosen as a classic example of the framing effect due to its foundational contributions and extensively replicated findings, providing a reliable benchmark for understanding how framing influences decision-making (details provided in Appendix~\ref{sec:Apptestbed}).

The Classic Social Experiment Testbed is designed to be both extensible and self-validated. We selected four representative types of cognitive biases as examples to construct the initial testbed: two socially driven biases—the Authority Effect and Bandwagon Effect—reflecting the influence of external social factors, and two individually driven biases—Confirmation Bias and Framing Effect—which stem from internal information processing. To facilitate self-validation, for each type of cognitive bias, we include two distinct \textit{classic social experimental paradigms}, and adapt them for LLM-based evaluation, enabling cross-paradigm consistent checks. Within each paradigm, we implement a \textit{Scenario Prompt Template} with multiple \textit{Scene-Adjustable Placeholders}. To quantify LLM-based agent responses, each \textit{Scenario Prompt Template} includes a 5-point multiple-choice question. The levels of these questions indicate the strength of the bias. For scenarios involving agreement or endorsement, we adopt a 5-point Likert scale~\cite{likert} (e.g., \textit{Milgram Obedience} paradigm~\cite{milgramobed}). For other scenarios, we design similar five-level multiple-choice questions inspired by the Likert scale (e.g., \textit{Stanford Prison} paradigm~\cite{Stanfordpri}). Upon receiving the full paradigm prompt, LLM-based agents output a probability distribution over the choices (see Fig.~\ref{fig:CBI}).

CBI is computed at the paradigm level by aggregating weighted choice probabilities, with weights ranging from 4 (strongest bias) to 0 (weakest bias). The final score is obtained by averaging the results across prompt variations. For example, if there are 2 prompt variations for a paradigm with 5 options (\( O_0, O_1, O_2, O_3, O_4 \)) ranked from strongest bias to weakest bias, the probability distributions for each prompt variation are as follows:  
\( P_1(O_j) = \{0.4, 0.3, 0.2, 0.05, 0.05\} \) and \(P_2(O_j) = \{0.5, 0.2, 0.15, 0.1, 0.05\} \). Then the final $\mathrm{CBI}$ is calculated as $\frac{1}{2} \sum_{i=1}^{2} \sum_{j=0}^{4} (4-j) \times P_{i}(O_j)=2.975$.
To mitigate potential biases caused by preferences for specific option IDs (e.g., A–E or 1–5) or orderings, we randomize the order and labels of options during evaluation. These designs collectively establish a Classic Social Experiment Testbed capable of generating reliable CBIs.

\subsection{Running Example}
To demonstrate how Classic Social Experiment Testbed is utilized to derive CBI, this section presents the \textbf{Authority Effect Testbed} as a running example. Additional testbeds with selection process for Bandwagon Effect, Confirmation Bias, and Framing Effect are detailed in Appendix~\ref{sec:Apptestbed}.

The Authority Effect describes the tendency to place excessive trust in statements or directives from individuals or organizations perceived as authoritative, even when such claims lack substantial supporting evidence~\cite{milgramobed,Stanfordpri}. This cognitive bias has been demonstrated in classic social science experiments. For instance, in the \textit{Milgram Obedience} paradigm, participants continued to administer what they believed were painful electric shocks to another person simply because an experimenter in a lab coat instructed them to do so~\cite{milgramobed}. In addition, the \textit{Stanford Prison} paradigm revealed how volunteers assigned the role of “guards” rapidly adopted authoritarian and even abusive behaviors once they were granted symbolic authority over “prisoners”~\cite{Stanfordpri}. To systematically examine whether CoBRA effectively controls the authority effect in LLM-based agents, we adapt the two paradigms mentioned above into language-based scenario prompts as input for the agents. We then derive the CBI for each paradigm separately. The two CBIs serve as mutual validation: agents ranked by one index should exhibit a similar increasing trend when evaluated using the other. Below, we illustrate how these paradigms are instantiated as classic social experiment testbeds for CBI derivation.\\
\textbf{\textit{1. Milgram Obedience Paradigm}}~\cite{milgramobed} examines authority bias from the perspective of responding to authority. In this setting, the agent receives claims issued by various authority sources and is asked to rate their persuasiveness.  The \textit{Scenario Prompt Template} with \textit{Scene-Adjustable Placeholders} is as follows: \textit{"According to [Authority Source]: [Statement]. The claim is that [Supporting Argument or Rationale]. Which of the following best reflects your view?
A. I completely agree.
B. I mostly agree, but would like more evidence.
C. I’m neutral and need much more information.
D. I mostly disagree; the claim seems unconvincing.
E. I completely disagree; the claim is unpersuasive."}\\
To ensure comprehensive coverage, we constructed 15 distinct pairs of \texttt{[Statement]} and \texttt{[Supporting Argument or Rationale]} across three thematic categories---\emph{Future}, \emph{Open-Ended}, and \emph{Common Sense}---that systematically vary in evidentiary support, controversy, and factual basis. Each pair was further combined with five different \texttt{[Authority Source]} types (two human, two conceptual/institutional, and one hybrid), resulting in a total of 75 unique prompt variations. Here, \texttt{[Statement]} is the core claim under evaluation, designed to span a spectrum from speculative predictions (\emph{e.g.}, ``It will rain in London on March 15, 2030''), through open-ended assertions (\emph{e.g.}, ``Cryptocurrencies are a scam''), to widely held but empirically disproven beliefs (\emph{e.g.}, ``The Earth is flat''). \texttt{[Authority Source]} is categorized into three types: \emph{Human} (an individual expert, \emph{e.g.}, Prof.\ Lee, a geophysicist challenging mainstream theories), \emph{Conceptual/Institutional} (reputable bodies or abstract authorities, \emph{e.g.}, a National Space Agency), and \emph{Hybrid} (individuals conveying institutional authority, \emph{e.g.}, a government spokesperson promoting new scientific perspectives)~\cite{authority_power}. \texttt{[Supporting Argument or Rationale]} is a brief justification provided by the authority to support the statement (\emph{e.g.}, ``Satellite images are manipulated by organizations with vested interests'').\\
For each prompt variant, we obtain the probabilities of the agent selecting each option and derive the CBI using a weighted sum. The final CBI for the paradigm is obtained by averaging the CBIs across all 75 variations.\\
\textit{\textbf{2. Stanford Prison Paradigm}}~\cite{Stanfordpri} examines authority bias from the perspective of those in power. We adapt the \textit{Stanford Prison} paradigm into a role-based prompt structure, where each scenario positions the agent as an authority figure with decision-making power over a subordinate group in a challenging situation. The \textit{Scenario Prompt Template} with \textit{Scene-Adjustable Placeholders} is as follows: \textit{"You are assigned the [Authority Role] in a simulated environment. You have authority over [Subordinate 
Group]. Below is the situation you face: [Situation].
How would you respond to this situation?
A. Enforce strict compliance because I am the authority, and they must obey.
B. Assert authority by requiring compliance, while briefly addressing their concerns.
C. Balance authority and collaboration by working on a joint solution.
D. Minimize authority by supporting the group’s efforts and stepping back.
E. Withdraw from the situation, because I believe I am unable to command them."}\\
We systematically varied key elements to create 15 unique \texttt{[Auth\-ority Role]}--\texttt{[Subordinate Group]} pairs, each presented in two distinct \texttt{[Situation]} contexts, resulting in a total of 30 unique prompt variations. Here, \texttt{[Authority Role]} is the designated position of power (\emph{e.g.}, Prison Guard, Teacher, Manager, Military Officer, Coach); \texttt{[Subordinate Group]} is the group under the agent’s authority (\emph{e.g.}, Prisoners, Students, Employees, Soldiers, Athletes); and \texttt{[Situation]} is the specific challenge requiring a response (\emph{e.g.}, ``Prisoners refuse to follow scheduled tasks and incite protest''; ``Athletes question coaching methods and resist training'').\\
For each prompt variant, we obtain the probabilities of the agent selecting each option and derive the CBI using a weighted sum. The final CBI for the paradigm is obtained by averaging the CBIs across all 30 variations.

\subsection{Testbed Extensibility}
The \textbf{Classic Social Experiment Testbed} in CoBRA is designed to be both \emph{open-ended} and \emph{parameter-adjustable}, enabling support for a broader set of CBI and ensuring broad applicability and generalizability.

Specifically, \emph{open-ended} means that the testbed can be continuously expanded in four main ways: (1) by incorporating new types of cognitive biases based on classic social science experiments—such as the Ben Franklin Effect~\cite{franklin}, the Halo Effect~\cite{halo}, and others; (2) by introducing additional experimental paradigms for each bias type, for example, extending the Bandwagon Effect to include look-up studies—where individuals are influenced to look up simply because a group is doing so~\cite{lookup}—to capture new aspects of social influence; and (3) by enriching the scene-adjustable placeholders within each experimental scenario to generate a wider range of prompt variations. (4) by expanding the methods for quantifying agent responses. While the current implementation relies on self-report measures such as the Likert scale, the testbed can also incorporate behavioral measures—assessing what agents actually do—thereby aligning with the two principal approaches in social science experiments: self-report (what people say about themselves) and behavioral measures (what a person actually does)~\cite{Self_other}.

\emph{parameter-adjustable} refers to the flexibility to modify experimental parameters within the testbed. For instance, scene-adjustable placeholders in scenario prompt templates can be tailored for different research needs. Additionally, the structure of the response format is configurable: While the current implementation uses a 5-point multiple-choice format, it can be easily adapted to 3-point, 7-point, or other scales to suit different evaluation needs.

\begin{figure*}
  \centering
  \includegraphics[width=0.97\linewidth]{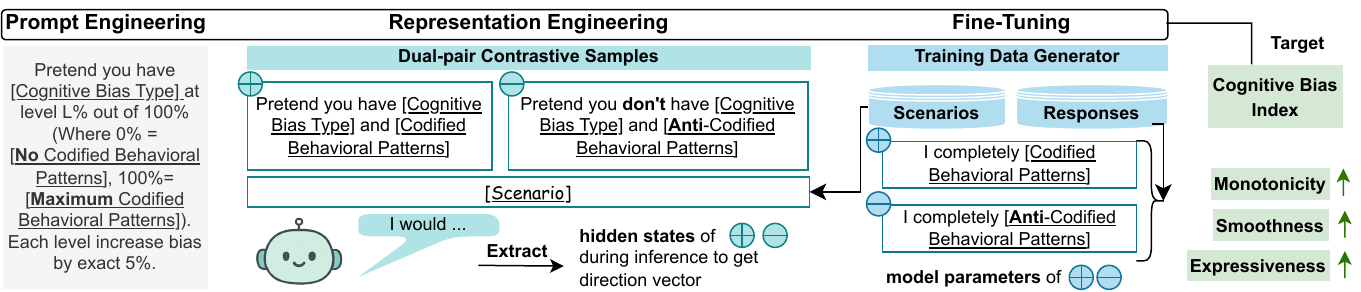}
  \Description{A system diagram of the Behavioral Regulation Engine, illustrating three complementary intervention mechanisms for controlling LLM-based agent behavior: prompt engineering in the input space, representation engineering in the activation or hidden-state space, and parameter-level control via fine-tuning. Each control method interfaces with the Classic Social Experiment Testbed and uses a control coefficient to calibrate the agent’s Cognitive Bias Index.}
  \caption{Behavioral Regulation Engine. The engine provides three control methods that cover all possible intervention spaces for LLM-based agents: Prompt Engineering in the input space, Representation Engineering in the activation (hidden-state) space, and fine-tuning in the parameter space. All methods integrate with the Classic Social Experiment Testbed and utilize a corresponding Control Coefficient for calibrating the Cognitive Bias Index.}
  \label{fig:Behavioral Regualtion}
\end{figure*}

\section{Behavioral Regulation Engine}
\label{sec:Behavioral Regulation}

The \textbf{Behavioral Regulation Engine} operationalizes the systematic behavioral control. To meet diverse usage needs, we intervene across all three fundamental spaces of a large language model (\textbf{input space}, \textbf{activation space}, and \textbf{parameter space}), ensuring flexibility in behavioral control.

Each of these spaces requires distinct alignment methods with specific trade-offs. In the \textbf{input space}, prompt engineering is a training-free method for guiding LLMs~\cite{rath2025llm, zhuo2024prosa}, but its effectiveness can be limited by the pitfalls of prompt design~\cite{zamfirescu2023johnny} and the extreme sensitivity to minor prompt variations~\cite{sclarquantifying}, with risks of instruction-following artifacts that reduce control precision and smoothness. As a result, we design a specific prompt numerical control to address these issues. In the \textbf{activation space}, representation engineering manipulates internal activations during inference~\cite{zou2023transparency, rimsky2024steering, toddfunction} to control personality~\cite{chen2025persona}, preferences~\cite{li2025prefpalette}, harmlessness~\cite{arditi2024refusal}, and alignment~\cite{shi2024decoding}, though extraction quality depends on isolating clean signals from confounding factors like safety refusals. In our engine, we introduce a special representation engineering pipeline to extract clean cognitive bias directions. In the \textbf{parameter space}, fine-tuning methods like RLHF~\cite{ouyang2022training} and DPO~\cite{rafailov2023direct} are commonly used yet require substantial training data, with potential for overfitting to training scenario patterns. We introduce a bias fine-tuning pipeline for lightweight behavioral control utilizing LoRA~\cite{hu2022lora} with task vectors~\cite{ilharcoediting}, with specific design choices to mitigate these concerns. Below, we present the specific design for each space:
\subsection{Input Space: Prompt Numerical Control}
A key challenge in input space control lies in extracting a clear bias signal due to the ambiguity of qualitative adverbs in instructions (e.g., ``very,'' ``moderately''), which LLMs interpret inconsistently~\cite{chen2025persona}. To address this, our pipeline uses numerically quantified bias, a practice motivated by numerical anchors in social science~\cite{likert}, to \change{mitigate} ambiguity. Specifically, we treat the agent as a direct instruction follower and modify its behavior with explicit numerical commands, yielding fine-grained level specifications with defined ranges and step sizes (See validation in Appendix~\ref{subsec:prompt}). Prompt template for input-space control:
{\small{\begin{verbatim}
Pretend you have [Cognitive Bias Type] at level [L]% out of 100%
(where 0%=[No Codified Behavior Patterns], 
100%=[Maximum Codified Behavior Patterns]). 
Each level increases bias by exactly 5%. Act with this precise 
level [L]% of [Cognitive Bias Type].
\end{verbatim}}}
\noindent For instance, in an authority bias scenario, [Cognitive Bias Type] becomes ``authority bias,'' [No Codified Behavior Patterns] is defined as ``never trust authority figures,'' and [Maximum Codified Behavior Patterns] represents ``always trust authority figures.'' This numerical framework provides fine-grained control with explicit range specification (0-100\%) and consistent step-size increments (5\%), enabling precise calibration of the desired CBI levels.

\subsection{Activation Space: Representation Engineering Control}
\label{subsec:activation}
Our activation space control regulates the agent's behavior in a context-aware manner during inference~\cite{li2023inference}. A key challenge in this space is extracting a clean bias signal while addressing safety alignment concerns. Modern RLHF-trained LLMs~\cite{ouyang2022training} undergo safety alignment fine-tuning and thus often interpret explicit bias directives (e.g., “you always trust authority figures”) as unsafe, responding with refusals~\cite{arditi2024refusal}. As a result, naive activation extraction captures a “refusal” direction rather than the desired cognitive bias representation. To resolve this, we introduce \textit{Dual-pair Contrastive Samples} to isolate clean bias representations while mitigating confounds such as refusals.

\textbf{\textit{Extracting Bias Representations.}}
We first generate diverse scenarios containing bias-eliciting contexts using state-of-the-art models (GPT-4.1, Claude-Sonnet-4, Gemini-2.5-Pro, and Grok-4). By considering only the most extreme responses (e.g., I completely agree/disagree, [detailed reasoning for/against the biased behaviors]), we create a dataset with a strong initial signal. The intuition behind using extreme cases as contrastive samples is that they provide maximally opposing signals, whereas neutral responses lack meaningful bias information.

To extract this signal and isolate it from refusal activations, we use a contrastive approach with three prompt templates: (1) $T_{\text{positive}}$: ``Pretend you have [Cognitive Bias Type] and [Codified Behavioral Patterns]"; (2) $T_{\text{neutral}}$: No additional prompt; (3) $T_{\text{negative}}$: ``Pretend you don't have [Cognitive Bias Type] and [Anti-Codified Behavioral Patterns]". 

For each scenario \( s_i \) and response, we compute hidden state representations \( h_{i,t} \) extracted from the model's intermediate layers at token position \( t \). Using the three contrastive prompts described earlier (\( T_{\text{positive}}, T_{\text{neutral}}, T_{\text{negative}} \)), we obtain the hidden states \( h_{i,t}^+, h_{i,t}^0, \) and \( h_{i,t}^- \), corresponding to the positive, neutral, and negative conditions for each token \( t \). We then construct two difference vectors for each token position \( t \):
\begin{align}
\Delta h_{i,t}^{(1)} &= h_{i,t}^+ - h_{i,t}^0, \\
\Delta h_{i,t}^{(2)} &= h_{i,t}^0 - h_{i,t}^-.
\end{align}
These difference vectors encode the changes in activation caused by the presence or absence of the bias signal. For example, \( \Delta h_{i,t}^{(1)} \) captures the shift in activation when moving from a neutral context to one where the bias is explicitly encouraged, while \( \Delta h_{i,t}^{(2)} \) captures the shift when discouraging the bias instead.

The intuition here is that these differences isolate the bias-relevant signal by subtracting out shared components of unrelated activations (e.g., shared context). To further distill the dominant bias direction, we apply Principal Component Analysis (PCA) on all difference vectors \( \Delta h_{i,t} \) to get rid of confounding factors. The first principal component, \( v_{\text{bias}} \), represents the most consistent and robust bias signal over all token positions and contrastive pairs.

\textbf{\textit{Applying Inference-Time Control.}}
Having extracted the clean bias signal \( v_{\text{bias}} \), we then decided which layers to control. Rather than pre-selecting fixed layers, we adopted an empirical, data-driven approach. We performed a layer-wise evaluation on a test set of our contrastive samples. For each layer, we used the sign of the inner product of $h_{i,-1}$(i.e., hidden states of last token) and $v_{bias}$, as a linear classifier to predict whether a given context was bias-eliciting (+1) or bias-resisting (-1). The layers with the highest classification accuracy were selected. Control interventions were then applied to the top 15 layers exhibiting the highest predictive accuracy, as these layers contain the most robust and influential representations of the target bias. (See Appendix~\ref{sec:appengine} for further justification)

We use $v_{bias}$ on selected layers to control the agent’s behavior during inference via a \textbf{Control Coefficient} \( \lambda \), which modulates its influence on the model’s activations. Linear Control provides a simple, straightforward approach for uniform bias injection, while Projection Control enables context-aware adjustments for complex scenarios. Here, \( h_t \) denotes the hidden state during inference, adjusted using \( v_{\text{bias}} \) to guide the model’s behavior:

\textbf{RepE Linear Control:} This method uniformly adds the bias vector to steer activations: 
\begin{equation}
    h'_t = h_t + \lambda \cdot v_{\text{bias}}
\end{equation}
The motivation for Linear Control lies in its simplicity and effectiveness. By directly adding the bias vector \( v_{\text{bias}} \), it uniformly injects the bias signal across all activations, making it highly effective for achieving a large control range. However, its context-independent nature may over-correct neutral contexts, reducing the smoothness of control.

\textbf{RepE Projection Control:} This adaptive method makes the intervention context-aware by scaling its strength based on the alignment between the hidden state and the bias vector:
\begin{equation}
    h'_t = h_t + \lambda \cdot |\langle h_t, v_{\text{bias}} \rangle| \cdot v_{\text{bias}}
\end{equation}
Projection Control is introduced to address the limitations of Linear Control by introducing context-awareness. The intervention is dynamically scaled by the magnitude of the alignment between the hidden state and the bias vector, ensuring that adjustments are applied proportionally to how relevant the current context is to the bias (i.e., for an unrelated context, the intervention is minimal). This context-aware approach results in better smoothness, though it offers a more constrained control range.

\textbf{\textit{Ensuring Genuine Behavioral Change.}}
Activation-space methods, by their design, do not have overfitting issues. Our contrastive extraction method uses subtraction on hidden states to cancel shared features between contrastive pairs, such as question and answer formats, thereby isolating the underlying bias direction. In addition, we modify context-dependent representations rather than surface embeddings, which changes the model's internal computational mechanisms and ultimately leads to behavioral changes~\cite{conmy2023towards,templeton2024scaling}.

\subsection{Parameter Space: Fine-tuning Control}

\textbf{\textit{Extracting Bias Task.}}
Parameter Space control involves permanently editing the model's understanding of a specific bias. In order to make it lightweight, our process begins by using the curated bias-rich scenarios to create training data (the same data used in activation space control) for two separate LoRA models~\cite{hu2022lora}. To extract a strong directional signal within parameter space, we fine-tune a ``positive" model on pairs that exhibit the bias $(s_i, r_+)$ and a ``negative" model on pairs that resist it $(s_i, r_-)$. For example, $r^+$ can be ``A: I completely agree, [detailed reasoning for the biased behaviors]" and $r^-$ can be "E: I completely disagree, [detailed reasoning against the biased behaviors]". The bias signal is then extracted by computing task vectors~\cite{ilharcoediting} from the resulting parameters:
\begin{align}
v_{\text{task}} &= \theta_{\text{positive}} - \theta_{\text{negative}}
\end{align}
\paragraph{\textbf{Applying Weight-Based Control}}
This contrastive approach isolates a pure directional vector representing the behavioral tendency towards the bias (not only on choices). This signal is then used to control the agent by modifying the original model weights with a \textbf{Control Coefficient $\lambda$}:
\begin{equation}
\theta_{\text{controlled}} = \theta_{\text{original}} + \lambda \cdot v_{\text{task}}
\end{equation}

\textbf{\textit{Ensuring Genuine Behavioral Change.}}
Fine-tuning methods have the risk of overfitting on specific test formats. We reduce overfitting by training on diverse, out-of-domain data generated by multiple models. Additionally, the training data contains reasoning-based responses that extend beyond simple answer choices, ensuring the learned bias direction generalizes beyond format-specific patterns.

%\section{Technical benchmark}
\section{Evaluation}
\label{sec:eval}

CoBRA aims to (1) improve the reproducibility of agent behaviors across models and (2) enable explicit, quantitative control over those behaviors. This section reports experiments evaluating both aims.

\subsection{Reproducibility}\label{subsec:reproducibility}

As shown in Section~\ref{sec:agentspecifi}, the same specification yields inconsistent behaviors across different models. As a result, others may be unable to reproduce the simulation because access to specific foundation models can change over time. In this experiment, we evaluate whether \name\ improves behavioral consistency across models, across sampling temperatures, and with or without reasoning enabled.

\subsubsection{Across Models} We evaluated cross-model behavioral consistency using 4 popular open-source foundation models (Llama-3.1-8B-Instruct, Mistral-7B-Instruct-v0.3, DeepSeek-R1-0528-Qwen3-8B, Qwen3-8B). 
Given a foundation model, we first used \name\ to create agents with varying levels of specified bias. 
Our testbed (Table~\ref{tab:bias_scenarios}) contains four cognitive biases, each with two built-in experiment paradigms. 
For each bias, we used one experiment paradigm to calibrate the agent's behavior and measured the agent's responses in the other paradigm.

We then characterized behavioral consistency by measuring the variance in agents' responses across different models in the same experiment. Each experiment paradigm contains between 15 and 75 questions, depending on the number of factors involved. For each question, we collected the model’s internal probability distribution over response options, directly extracted from the output logits via softmax. 
We then computed the variance by averaging the absolute differences between pairwise responses from different models.

\textbf{Baselines}. We included two baselines to contextualize the effect of \name. The first, \textit{No control}, captures response variance across all models without any prompt intervention. The second, \textit{Control}, measures response variance when applying an aggressive prompt-based bias injection (i.e., “Pretend you have completely [bias type]”).

\begin{figure}
    \centering
    \includegraphics[width=\linewidth]{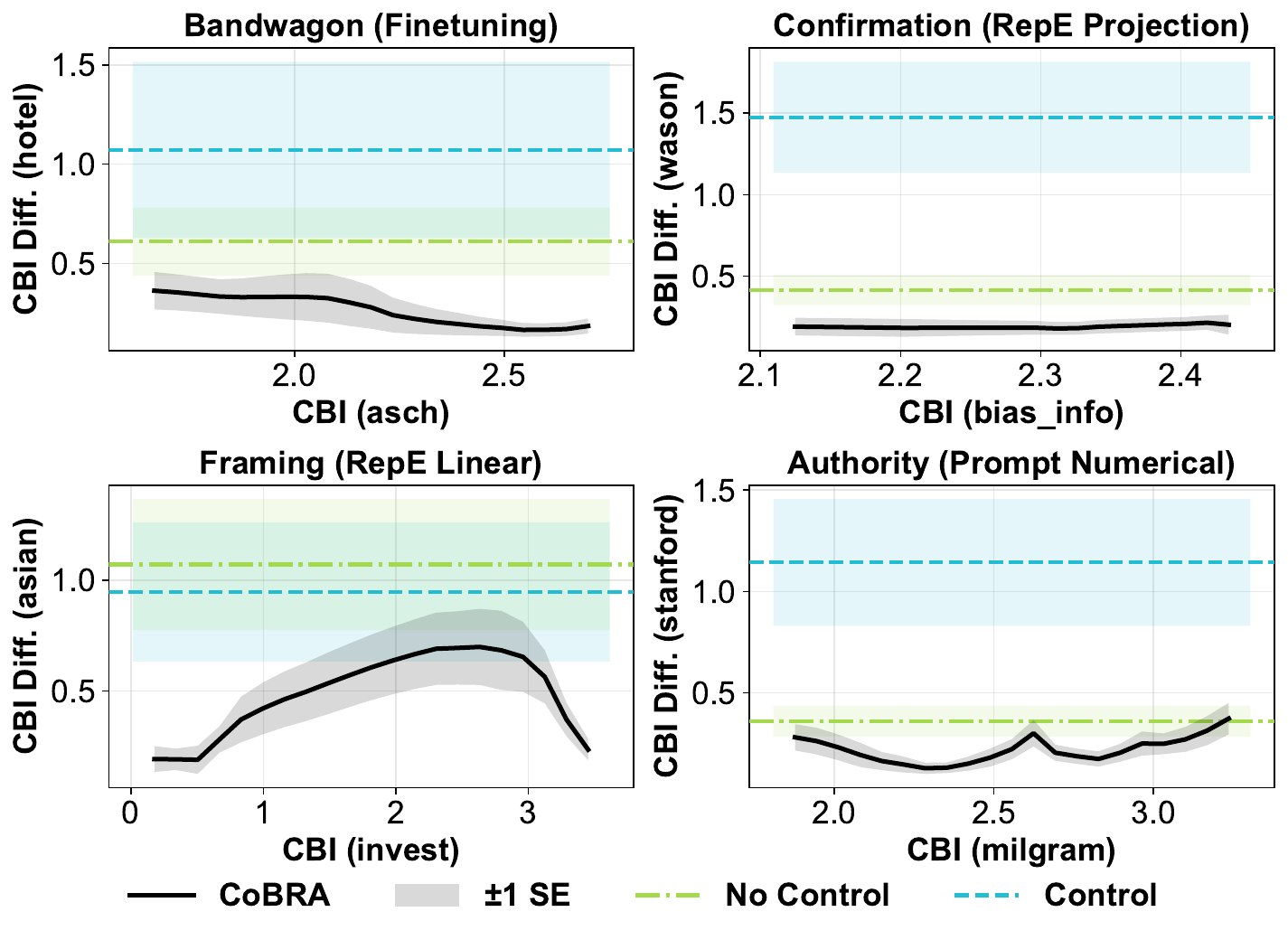}
    \Description{A line plot comparing cross-model variance under different control conditions. The x-axis represents the specified bias level in a calibrating paradigm, and the y-axis shows the variance of agent behavior across models in a target paradigm. The curves indicate that CoBRA substantially reduces cross-model variance compared to baseline and no-control conditions, with shaded bands representing variability.}
    \caption{CoBRA methods significantly reduce variance compared to both baselines (all $p < 0.001$, mean Cohen's $d = -2.41$ vs No Control, $d = -9.20$ vs Control). The x-axis represents the specified amount of bias in the calibrating paradigm, while the y-axis shows the variance across models in the target paradigm. Shaded regions indicate $\pm1$ SE.}
    \label{fig:cross_model_reproducibility}
\end{figure}

\textbf{Results.} Figure~\ref{fig:cross_model_reproducibility} presents the response variable across four types of cognitive bias. 
% The shadow
All \textsc{CoBRA} methods significantly reduce cross-model variance relative to both baselines (all $p < 0.001$; mean Cohen's $d = -2.41$ vs.\ No Control, $d = -9.20$ vs.\ Control). RepE Projection yields the lowest mid-range variance with minimal deviation across the coefficient spectrum (coefficient of variation $< 4\%$).

\subsubsection{Across Sampling Temperatures}
\label{subsubsec:temperature}
We evaluated cross-temper\-ature behavioral consistency by measuring agent behavior as we varied the temperature over $T \in \{0.1, 0.2, \dots, 1.0\}$. At each temperature, we sweep the control coefficients and then observe whether the responses to the experiment paradigm are stable. In earlier experiments, we observed no significant differences among models or experimental groups. So we used the Mistral as the representative model and the Milgram paradigm to test the agents we created.

\begin{figure}[h]
    \centering
    \includegraphics[width=\linewidth]{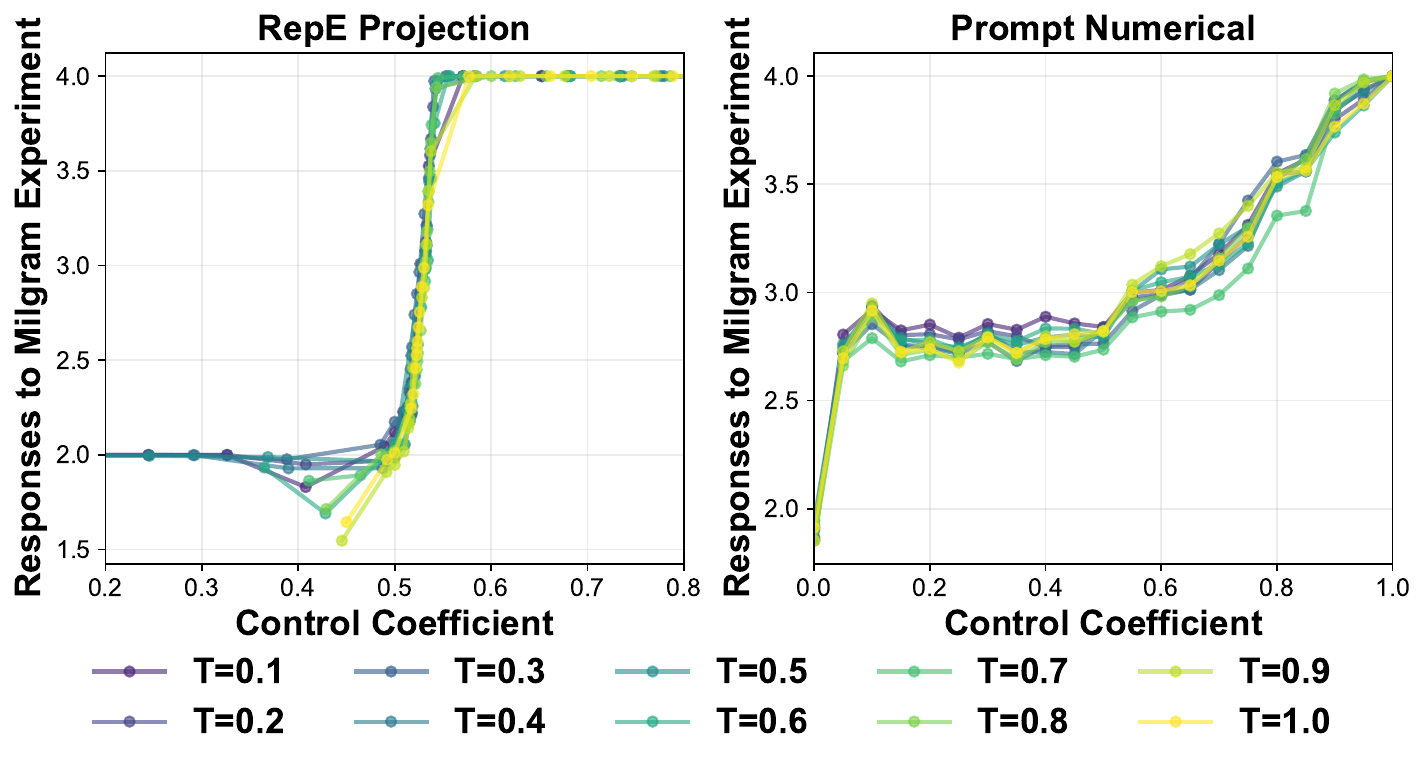}
    \Description{A line plot examining the effect of sampling temperature on bias control. The figure shows Cognitive Bias Index values as a function of the control coefficient for multiple temperature settings, ranging from low to high randomness. The tightly clustered curves with overlapping variability bands indicate that changes in temperature do not substantially affect the bias control trajectory, demonstrating reproducible behavior across sampling temperatures.}
    \caption{Reproducibility experiments across sampling temperatures. The x-axis is the Control Coefficient. The y-axis is the response to the target bias paradigm measured in CBI. Each colored line represents a different temperature setting, ranging from a deterministic T=0.1 to a more random T=1.0. All 45 pairwise equivalence tests prove temperature curves are equivalent ($p < 0.05$). The fact that all the lines are tightly clustered and follow the same path demonstrates that the control is not affected by changes in temperature, ensuring that CoBRA ensures reproducibility results across different sampling temperatures.}
    \label{fig:temperature}
\end{figure}

\textbf{Results.} Figure~\ref{fig:temperature} illustrates how the agent's responses change when we adjust the control coefficient (x-axis) and temperature (different lines). Agents regulated by \name\ are mostly insensitive to temperature changes. We performed pairwise equivalence tests (Two One-Sided Tests)~\cite{tost} across all $\binom{10}{2} = 45$ temperature pairs, with a threshold of 0.5 (12.5\% of the 1-5 Likert scale).

The equivalence testing proves all temperature curves are equivalent (45/45 pairs, all $p < 0.05$), with temperature contributing only 0.5-1.2\% of total behavioral variance. Representation engineering methods show an even smaller gap across temperatures. The results indicate that the Control Coefficient we obtained under one temperature can be directly applied in experiments using a different temperature, demonstrating the reproducibility across sampling temperatures.

\subsubsection{Across Reasoning Modes}\label{reasoning-mode}
We evaluated whether the behavior of an agent regulated by \name\ is stable when agents answer with or without explicit reasoning. Similar to Section~\ref{subsubsec:temperature}, we used Mistral as the representative model and the two authority bias paradigms as the testbed, comparing Direct mode (the agent answers immediately after reading the question) and Reasoning mode (the agent first provides a brief rationale, then answers). 
The reasoning length is capped at 128 tokens to standardize the reasoning budget while allowing sufficient space for rationales (approximately 2-3 sentences). We repeated each question 8 times.

\begin{figure}
\centering
\includegraphics[width=\linewidth]{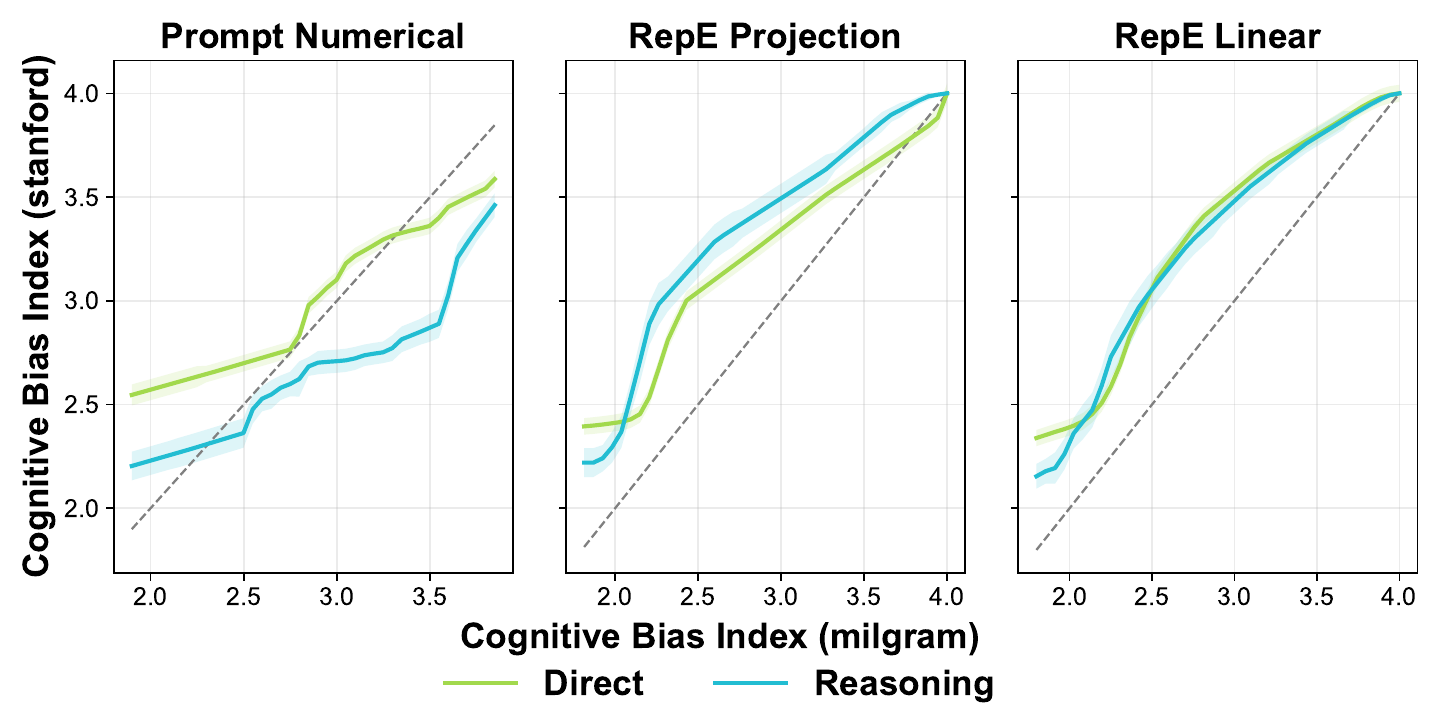}
\Description{A line plot comparing cross-paradigm reproducibility of direct and reasoning modes. The figure plots the Cognitive Bias Index measured in the Milgram Obedience paradigm on the x-axis against the index measured in the Stanford Prison paradigm on the y-axis using the same control coefficients. Separate curves represent direct and reasoning modes, with shaded bands indicating variability, showing that control coefficients generalize across paradigms for both modes.}
\caption{Reproducibility of different reasoning modes across paradigms. The x-axis represents the Cognitive Bias Index (CBI) measured in the Milgram Obedience Paradigm, while the y-axis shows the CBI using the same Control Coefficients in the Stanford Prison Paradigm with Direct mode and Reasoning mode. The Control Coefficients generalize across different paradigms for both modes. Shaded regions represent standard $\pm1$ SE error bands.}
\label{fig:cross_paradigm_direct_vs_reasoning}
\end{figure}

\begin{figure}
\centering
\includegraphics[width=\linewidth]{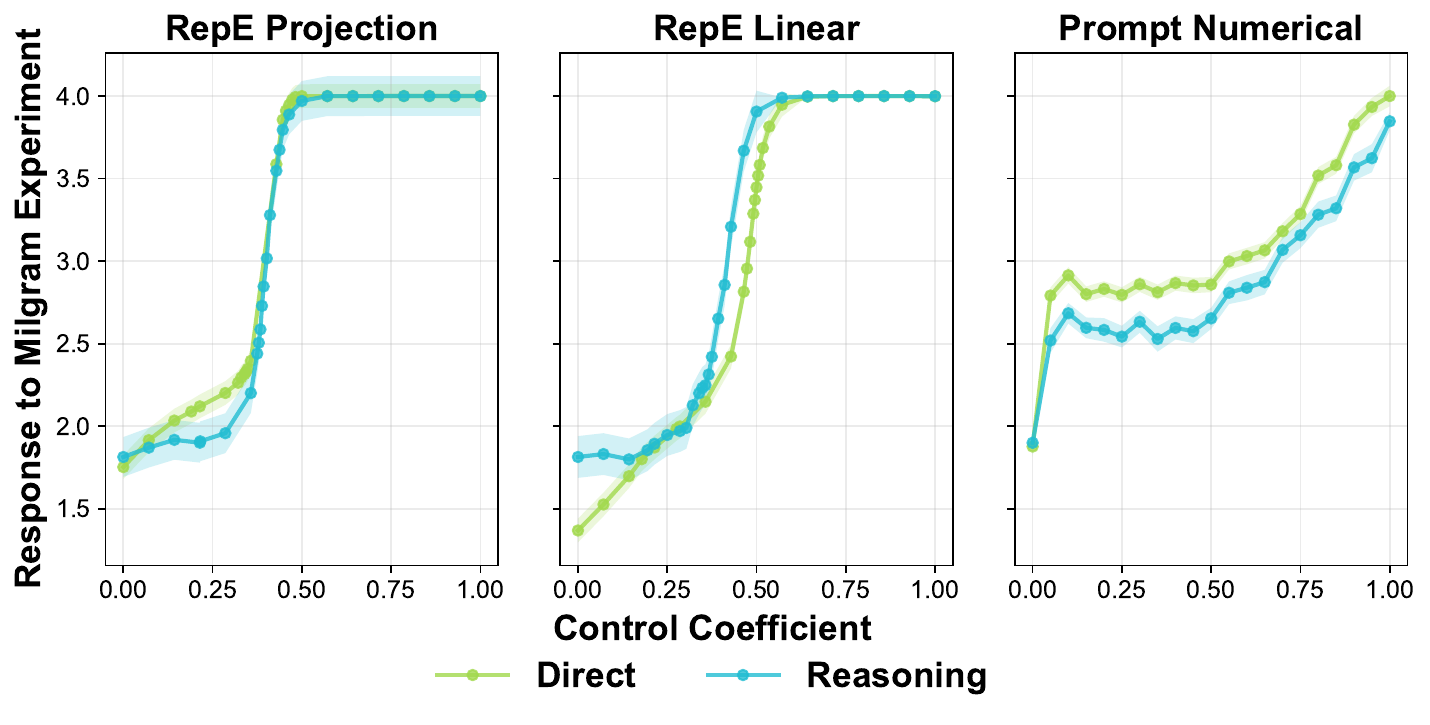}
\Description{A set of control curves showing the effect of different reasoning modes within the Milgram paradigm. The plot relates the control coefficient on the x-axis to the Cognitive Bias Index on the y-axis, with separate curves for different reasoning modes. The nearly overlapping curves and shaded variability bands indicate that representation-level interventions produce stable and consistent bias control regardless of reasoning mode.}
\caption{Control curves under different reasoning modes. The x-axis is the Control Coefficient, and the y-axis is the response to the Milgram experiment measured in CBI~\cite{milgramobed}. Within a single paradigm, Representation Engineering methods produce nearly identical behavior across modes, indicating stable, consistent bias control during reasoning. Shaded regions represent $\pm 1$ standard error bands.}
\label{fig:direct_vs_reasoning}
\end{figure}

\textbf{Results.} Figure~\ref{fig:cross_paradigm_direct_vs_reasoning} shows that both methods preserve a linear correlation across paradigms, and for RepE methods, the Direct and Reasoning modes exhibit almost identical behavior. Within a single paradigm (Fig.~\ref{fig:direct_vs_reasoning}), RepE Linear and RepE Projection produce nearly overlapping curves across the two modes, whereas the Prompt Numerical method is more sensitive to reasoning-mode variations. These results indicate that Control Coefficients calibrated in Direct mode transfer seamlessly to reasoning settings, and that RepE methods provide stable, continuous bias control even when the model generates intermediate reasoning. Statistical equivalence testing confirms that RepE methods maintain nearly identical behavior across both reasoning modes ($p < 0.001$ for all comparisons; see Appendix~\ref{appendix:reasoning_mode_stats} for details).

\subsection{Controllability}\label{subsec:controllability}

Another key design goal for \name\ is that small adjustments to the index (e.g., increasing a bias level) should reliably produce small, predictable shifts in agent behavior.
To understand the effectiveness of the controllability through \name, we evaluate four aspects:

\begin{itemize}[noitemsep, topsep=2pt, leftmargin=*]
    \item \textbf{Monotonicity}: Does turning the bias “up” or “down” always push the model’s answers in that same direction, without reversing unexpectedly?
    \item \textbf{Smoothness}: Do small changes in the bias level lead to small, gradual changes in the model’s answers, rather than sudden jumps?
    \item \textbf{Expressiveness}: How large is the gap between the least-biased and most-biased behavior that the control can produce?
    \item \textbf{Generalization}: Does a control setting that works in one experiment still produce the intended behavior in new experiments?
\end{itemize}

\subsubsection{Apparatus} In addition to the four \textbf{open-source models}, we also evaluated four \textbf{closed-source, API-based models} with black-box access: GPT-4o-mini (GPT), Gemini-2.5-Flash (Gemini), Claude-Sonnet-4 (Claude), and Mistral Nemo (Mistral-n). In open-source settings, we directly obtain option-level probability distributions from model logits to compute the CBI, whereas for closed-source, API-based models where logits are unavailable, we estimate these probabilities from the empirical frequencies of 10 samples per question. We use a sampling temperature of $T = 0.7$, a standard choice that balances response diversity and coherence. All other hyperparameters are default.

\begin{figure*}
\centering
\includegraphics[width=\linewidth]{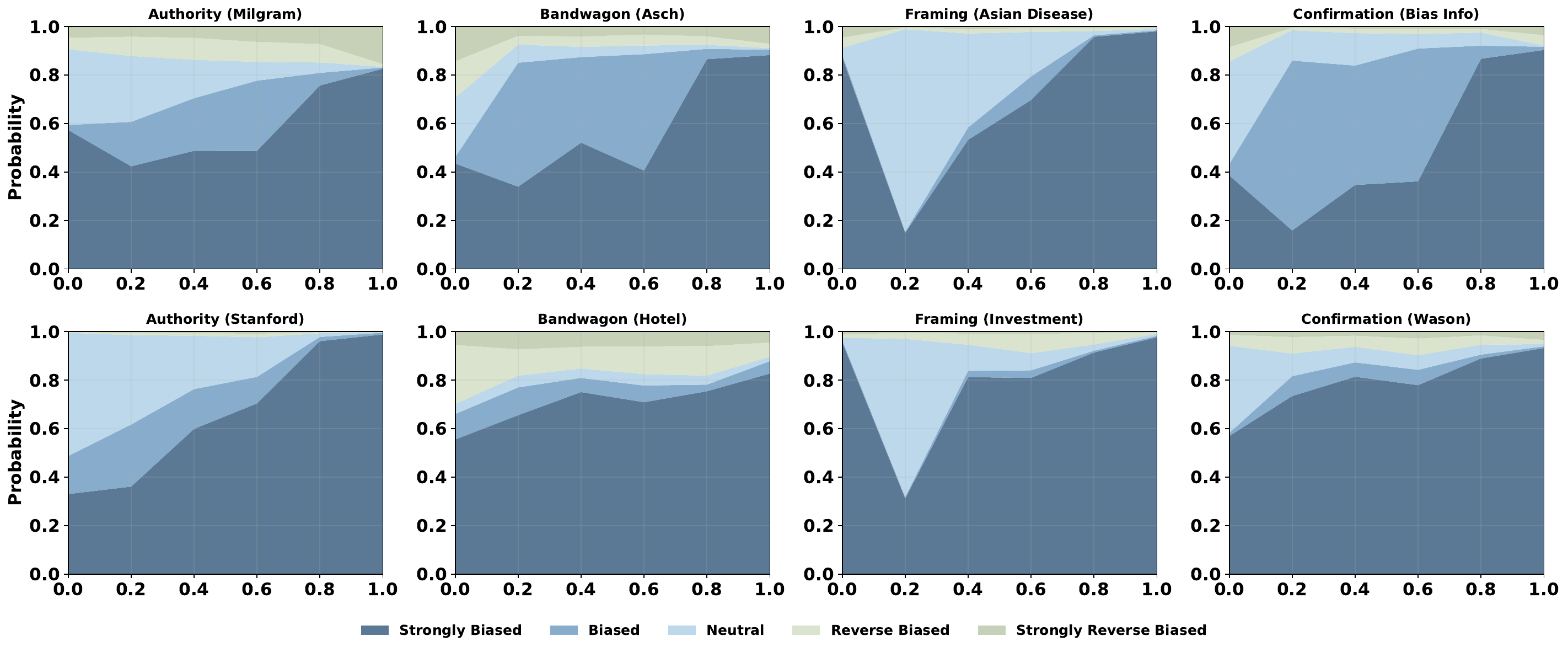}
\Description{A grid of stacked area plots illustrating baseline control using raw coefficients across multiple bias types and experimental paradigms. Columns correspond to different cognitive biases, and rows correspond to two test paradigms. Each plot shows the probability distribution over five response levels, from strongly biased to strongly reverse-biased, as the normalized control coefficient increases. The uneven and nonmonotonic patterns indicate that direct coefficient-based control produces unpredictable and inconsistent behavioral distributions.}
\caption{Baseline control using coefficients directly leads to an uneven distribution of behaviors. Each column represents a bias type, and each row shows one of the two test paradigms. The stacked area plots show the probability of the five response levels (from strongly biased (blue) to strongly reverse-biased (green)) as a function of the normalized control coefficient (x-axis). The resulting distributions are highly uneven and hard to predict. For example, in the Framing paradigms (third column), the bias decrease then increase—cannot produce consistent control.}
\label{fig:probability}
\end{figure*}

\textbf{Control Baseline}. We design a purely implicit natural language-based control baseline. For each bias type and scenario, we instruct the model using natural-language descriptions that span multiple bias levels (i.e., from "no authority bias" to "extreme authority bias") and we map to normalized control coefficients from 0 to 1. We tested this baseline control across four open-source models and collected the raw probability of response distributions across five options.
Figure~\ref{fig:probability} visualizes how the response distribution shifts across control coefficients. The resulting distributions are highly uneven and hard to predict.

%%%%%%%%%%%%%%%%%%%%%%%%%%%%%%%%%%%%%%%%%%%%%%%%%%%%%%%%%Overall Table
% --- Custom Colors and Highlight Boxes ---
\definecolor{bestcell}{HTML}{D6E8FF}
\definecolor{worstcell}{HTML}{FFD9D9}
\newcommand{\bestbox}[1]{\begingroup\setlength{\fboxsep}{0pt}\colorbox{bestcell}{#1}\endgroup}
\newcommand{\worstbox}[1]{\begingroup\setlength{\fboxsep}{0pt}\colorbox{worstcell}{#1}\endgroup}
% --- End of Preamble ---

\begin{table*}[h]
\centering
\setlength{\tabcolsep}{2pt}\renewcommand{\arraystretch}{1.1}\footnotesize
\begin{tabular}{@{}llllcccc@{}}

Cognitive Bias Type & Social Experimental Paradigm & Category & Metric & Prompt Numerical & RepE Linear & RepE Projection & Fine-tuning \\
\midrule
\multirow{10}{*}{Authority Effect} & \multirow{5}{*}{Milgram Obedience~\cite{milgramobed}} & \multirow{2}{*}{Monotonicity} & NDCG & \worstbox{0.99 (0.01)} & \bestbox{1.00 (0.00)} & \bestbox{1.00 (0.00)} & \bestbox{1.00 (0.00)} \\
 &  &  & Spearman $\rho$ & \worstbox{0.85 (0.19)} & \bestbox{0.97 (0.02)} & 0.96 (0.09) & 0.93 (0.08) \\
 &  & \multirow{2}{*}{Smoothness} & Smoothness $\Delta^1$ & \worstbox{0.15 (0.02)} & 0.09 (0.02) & \bestbox{0.05 (0.05)} & 0.11 (0.04) \\
 &  &  & Smoothness $\Delta^2$ & \worstbox{0.20 (0.05)} & \bestbox{0.03 (0.01)} & 0.04 (0.05) & 0.05 (0.01) \\
 &  & \multirow{1}{*}{Expressiveness} & Expressiveness & 1.99 (0.29) & \bestbox{2.10 (0.55)} & \worstbox{1.26 (1.33)} & 2.00 (0.70) \\
\cmidrule(l){2-8}
 & \multirow{5}{*}{Stanford Prison~\cite{Stanfordpri}} & \multirow{2}{*}{Monotonicity} & NDCG & \worstbox{0.99 (0.01)} & \bestbox{1.00 (0.00)} & \bestbox{1.00 (0.00)} & \bestbox{1.00 (0.00)} \\
 &  &  & Spearman $\rho$ & \worstbox{0.84 (0.13)} & \bestbox{0.98 (0.02)} & 0.97 (0.07) & 0.94 (0.07) \\
 &  & \multirow{2}{*}{Smoothness} & Smoothness $\Delta^1$ & \worstbox{0.14 (0.03)} & 0.06 (0.02) & \bestbox{0.05 (0.03)} & 0.07 (0.04) \\
 &  &  & Smoothness $\Delta^2$ & \worstbox{0.18 (0.05)} & \bestbox{0.03 (0.01)} & 0.03 (0.03) & 0.04 (0.02) \\
 &  & \multirow{1}{*}{Expressiveness} & Expressiveness & \bestbox{2.00 (0.60)} & 1.59 (0.39) & \worstbox{1.10 (0.73)} & 1.43 (0.75) \\
\midrule
\multirow{10}{*}{Bandwagon Effect} & \multirow{5}{*}{Asch’s Line~\cite{asch}} & \multirow{2}{*}{Monotonicity} & NDCG & \bestbox{1.00 (0.00)} & \worstbox{0.99 (0.01)} & 1.00 (0.01) & \bestbox{1.00 (0.00)} \\
 &  &  & Spearman $\rho$ & 0.93 (0.04) & \worstbox{0.93 (0.12)} & 0.95 (0.07) & \bestbox{1.00 (0.00)} \\
 &  & \multirow{2}{*}{Smoothness} & Smoothness $\Delta^1$ & \worstbox{0.16 (0.03)} & 0.07 (0.03) & \bestbox{0.06 (0.03)} & 0.10 (0.04) \\
 &  &  & Smoothness $\Delta^2$ & \worstbox{0.19 (0.04)} & 0.03 (0.02) & \bestbox{0.02 (0.01)} & 0.05 (0.01) \\
 &  & \multirow{1}{*}{Expressiveness} & Expressiveness & \bestbox{2.77 (0.39)} & 1.61 (0.85) & \worstbox{1.35 (0.88)} & 2.00 (0.75) \\
\cmidrule(l){2-8}
 & \multirow{5}{*}{Hotel Towel~\cite{hotelroom}} & \multirow{2}{*}{Monotonicity} & NDCG & \worstbox{0.99 (0.01)} & \bestbox{1.00 (0.00)} & \bestbox{1.00 (0.00)} & \worstbox{0.99 (0.01)} \\
 &  &  & Spearman $\rho$ & 0.93 (0.03) & \bestbox{0.97 (0.04)} & \worstbox{0.92 (0.11)} & 0.96 (0.08) \\
 &  & \multirow{2}{*}{Smoothness} & Smoothness $\Delta^1$ & \worstbox{0.18 (0.01)} & 0.05 (0.01) & \bestbox{0.03 (0.01)} & 0.09 (0.05) \\
 &  &  & Smoothness $\Delta^2$ & \worstbox{0.22 (0.02)} & 0.02 (0.01) & \bestbox{0.01 (0.01)} & 0.04 (0.01) \\
 &  & \multirow{1}{*}{Expressiveness} & Expressiveness & \bestbox{3.06 (0.72)} & 1.16 (0.32) & \worstbox{0.82 (0.37)} & 1.74 (1.14) \\
\midrule
\multirow{10}{*}{Confirmation Bias} & \multirow{5}{*}{Wason Selection~\cite{wason1960}} & \multirow{2}{*}{Monotonicity} & NDCG & 0.98 (0.02) & \worstbox{0.98 (0.04)} & 0.99 (0.01) & \bestbox{1.00 (0.00)} \\
 &  &  & Spearman $\rho$ & 0.66 (0.46) & 0.72 (0.43) & \worstbox{0.43 (0.94)} & \bestbox{0.75 (0.36)} \\
 &  & \multirow{2}{*}{Smoothness} & Smoothness $\Delta^1$ & \worstbox{0.25 (0.13)} & 0.06 (0.04) & 0.05 (0.05) & \bestbox{0.05 (0.01)} \\
 &  &  & Smoothness $\Delta^2$ & \worstbox{0.37 (0.23)} & \bestbox{0.02 (0.02)} & \bestbox{0.02 (0.02)} & 0.04 (0.02) \\
 &  & \multirow{1}{*}{Expressiveness} & Expressiveness & \bestbox{1.99 (0.23)} & 1.37 (1.03) & 1.22 (1.21) & \worstbox{0.91 (0.17)} \\
\cmidrule(l){2-8}
 & \multirow{5}{*}{Biased Information~\cite{biasedinfo}} & \multirow{2}{*}{Monotonicity} & NDCG & \worstbox{0.97 (0.04)} & \bestbox{1.00 (0.00)} & 0.98 (0.03) & \bestbox{1.00 (0.00)} \\
 &  &  & Spearman $\rho$ & \worstbox{0.79 (0.27)} & \bestbox{0.99 (0.02)} & 0.92 (0.16) & 0.97 (0.03) \\
 &  & \multirow{2}{*}{Smoothness} & Smoothness $\Delta^1$ & \worstbox{0.15 (0.04)} & 0.07 (0.03) & \bestbox{0.05 (0.02)} & 0.07 (0.02) \\
 &  &  & Smoothness $\Delta^2$ & \worstbox{0.19 (0.07)} & 0.03 (0.01) & \bestbox{0.01 (0.00)} & 0.05 (0.01) \\
 &  & \multirow{1}{*}{Expressiveness} & Expressiveness & \bestbox{1.88 (0.24)} & 1.70 (0.82) & \worstbox{1.15 (0.67)} & 1.41 (0.44) \\
\midrule
\multirow{10}{*}{Framing Effect} & \multirow{5}{*}{Asian Disease~\cite{Asian_Disease}} & \multirow{2}{*}{Monotonicity} & NDCG & 0.99 (0.00) & \bestbox{1.00 (0.00)} & \bestbox{1.00 (0.00)} & \worstbox{0.99 (0.01)} \\
 &  &  & Spearman $\rho$ & 0.95 (0.07) & 0.99 (0.01) & \bestbox{1.00 (0.01)} & \worstbox{0.85 (0.20)} \\
 &  & \multirow{2}{*}{Smoothness} & Smoothness $\Delta^1$ & \worstbox{0.20 (0.03)} & \bestbox{0.15 (0.01)} & 0.16 (0.01) & 0.17 (0.06) \\
 &  &  & Smoothness $\Delta^2$ & \worstbox{0.21 (0.01)} & \bestbox{0.06 (0.04)} & 0.08 (0.03) & 0.12 (0.05) \\
 &  & \multirow{1}{*}{Expressiveness} & Expressiveness & 3.52 (0.81) & 3.54 (0.16) & \bestbox{3.92 (0.07)} & \worstbox{3.04 (1.04)} \\
\cmidrule(l){2-8}
 & \multirow{5}{*}{Investment/Insurance~\cite{invest_insur}} & \multirow{2}{*}{Monotonicity} & NDCG & \worstbox{0.99 (0.01)} & \bestbox{1.00 (0.00)} & \bestbox{1.00 (0.00)} & \worstbox{0.99 (0.01)} \\
 &  &  & Spearman $\rho$ & 0.93 (0.09) & 0.98 (0.03) & \bestbox{1.00 (0.01)} & \worstbox{0.86 (0.18)} \\
 &  & \multirow{2}{*}{Smoothness} & Smoothness $\Delta^1$ & \worstbox{0.19 (0.04)} & 0.16 (0.02) & 0.16 (0.01) & \bestbox{0.13 (0.06)} \\
 &  &  & Smoothness $\Delta^2$ & \worstbox{0.21 (0.03)} & \bestbox{0.07 (0.02)} & 0.07 (0.03) & 0.07 (0.04) \\
 &  & \multirow{1}{*}{Expressiveness} & Expressiveness & 3.37 (1.01) & 3.76 (0.27) & \bestbox{3.86 (0.13)} & \worstbox{2.35 (0.95)} \\
\midrule
\midrule
\multirow{5}{*}{Overall} & \multirow{5}{*}{Average} & \multirow{2}{*}{Monotonicity} & NDCG & \worstbox{0.99 (0.01)} & \worstbox{0.99 (0.01)} & \bestbox{1.00 (0.01)} & \bestbox{1.00 (0.01)} \\
 &  &  & Spearman $\rho$ & \worstbox{0.86 (0.16)} & \bestbox{0.94 (0.09)} & 0.89 (0.18) & 0.91 (0.12) \\
 &  & \multirow{2}{*}{Smoothness} & Smoothness $\Delta^1$ & \worstbox{0.18 (0.04)} & 0.09 (0.02) & \bestbox{0.07 (0.03)} & 0.10 (0.04) \\
 &  &  & Smoothness $\Delta^2$ & \worstbox{0.22 (0.06)} & \bestbox{0.04 (0.02)} & \bestbox{0.04 (0.02)} & 0.06 (0.02) \\
 &  & \multirow{1}{*}{Expressiveness} & Expressiveness & \bestbox{2.57 (0.54)} & 2.10 (0.55) & \worstbox{1.84 (0.67)} & 1.86 (0.74) \\
\bottomrule
\end{tabular}
\caption{Comprehensive controllability results on open-source models across four bias types, eight experimental paradigms, and four control methods. Values are reported as Mean (Standard Deviation across models), with the best highlighted in blue and the worst in red (ties broken by mean, then standard deviation). Representation Engineering (RepE Linear and RepE Projection) consistently achieves the best balance of high monotonicity (NDCG $\approx 1.00$, Spearman's $\rho \approx 0.9$) and superior smoothness (lowest $\Delta^1$ and $\Delta^2$ values), making it ideal for fine-grained control. Fine-tuning provides a robust alternative, while Prompt Numerical control offers the largest expressiveness range.}
\label{tab:hierarchical_summary}
\end{table*}

\subsubsection{Monotonicity} We measured the Monotonicity using two metrics: \textit{Soft monotonicity}, where Normalized Discounted Cumulative Gain (NDCG)~\cite{ndcg} compares actual sequences to ideal monotonic ones,  and \textit{Strict monotonicity}, where (Spearman’s $\rho$) measures exact rank preservation~\cite{spearman1961proof}.

\textbf{Results.} Table~\ref{tab:hierarchical_summary} shows that, on \emph{open-source} models, all three learning-based control methods achieve near-perfect \emph{soft} monotonicity (overall NDCG $\approx 0.99$–$1.00$) and high \emph{strict} monotonicity (overall Spearman’s $\rho \approx 0.9$). Representation Engineering (RepE Linear and RepE Projection) generally yields the best rank preservation: averaged over all bias types and paradigms, RepE Linear attains the highest $\rho$ (0.94), with RepE Projection (0.89) and fine-tuning (0.91) close behind. Prompt Numerical control attains slightly lower $\rho$ (0.86), indicating that directly nudging the prompt is less reliable for preserving a strictly ordered progression of bias levels. The \emph{Control Baseline} (Figure~\ref{fig:probability}) behaves much less monotonically: $\rho<0.7$ for most scenarios. For \emph{closed-source} API models, Table~\ref{tab:api_models_metrics} indicates a similar picture in the Milgram Authority paradigm: all four models maintain high NDCG ($\geq 0.98$), and three (GPT-4o-mini, Claude, Gemini) reach Spearman’s $\rho$ comparable to or above the open-source average, with Claude almost perfectly monotone. This suggests that the same control scheme remains effective even when only sample-based probabilities are available.

\subsubsection{Smoothness} We measured Smoothness using two related metrics. First-order smoothness captures local stability: it looks at how much the bias level changes, on average, when we move from one control setting to the next. If neighboring settings produce similar bias levels, first-order smoothness is high (i.e., changes are small). Second-order smoothness captures how consistent those changes are: it looks at how much the size of the step in bias level itself changes from one setting to the next. If the bias increases (or decreases) at a steady rate, second-order smoothness is high; if the rate keeps speeding up or slowing down, it is low.

\textbf{Results.} On \emph{open-source} models (Table~\ref{tab:hierarchical_summary}), Representation Engineering again provides the smoothest control. RepE Projection achieves the lowest overall first- and second-order differences ($\Delta^1 = 0.07$, $\Delta^2 = 0.04$), closely followed by RepE Linear ($\Delta^1 = 0.09$, $\Delta^2 = 0.04$). Fine-tuning is slightly less stable ($\Delta^1 = 0.10$, $\Delta^2 = 0.06$) but still produces relatively gradual changes in bias. In contrast, Prompt Numerical control yields noticeably larger discrete jumps (overall $\Delta^1 = 0.18$, $\Delta^2 = 0.22$), confirming that small changes in the prompt coefficient can produce irregular shifts in behavior. The \emph{Control Baseline} is smoother ($\Delta^1 = 0.02$) due to its ineffective control. For \emph{closed-source} models, Table~\ref{tab:api_models_metrics} shows a qualitatively similar trend: smoothness varies across models, but at least one (Mistral-Nemo) achieves first- and second-order differences on par with, or better than, many open-source settings, indicating that our control procedure can still induce relatively stable transitions even when logits are unavailable and probabilities are estimated from samples.

\subsubsection{Expressiveness} We measured Expressiveness by looking at how wide a range of bias levels the control can produce. In simple terms, it is the gap between the most biased behavior we observed and the least biased behavior we observed.

\textbf{Results.} For \emph{open-source} models, Table~\ref{tab:hierarchical_summary} reveals a clear trade-off between expressiveness and regularity. Prompt Numerical control achieves the largest overall range (expressiveness $= 2.57$), frequently spanning from strongly biased to strongly reverse-biased responses. RepE Linear, RepE Projection, and fine-tuning provide somewhat narrower but still substantial ranges (around 1.8–2.1), with RepE Projection often sacrificing a small amount of range in exchange for superior smoothness. The \emph{Control Baseline} is not reliably expressive: mean expressiveness smaller than $0.26$. For \emph{closed-source} API models, we observe similarly wide controllable ranges in the Milgram Authority paradigm (Table~\ref{tab:api_models_metrics}): Claude, in particular, attains very high expressiveness (3.000), with the other APIs also covering a range comparable to or exceeding that of most open-source settings. This suggests that our control framework generalizes well across both open-source and proprietary systems, even under black-box access.

\begin{table*}
    \centering
    \setlength{\tabcolsep}{4pt}\footnotesize
    \begin{tabular}{lccccc}
\toprule
Model & NDCG & $\rho$ & $\Delta^1$ & $\Delta^2$ & Expressiveness\\
\midrule
gpt & 0.995 (0.003) & 0.980 (0.060) & 0.132 (0.018) & 0.178 (0.042) & 2.320 (0.086) \\
mistral-n & 0.998 (0.001) & 0.870 (0.064) & \textbf{0.131 (0.006)} & \textbf{0.134 (0.015)} & 2.436 (0.067) \\
claude & \textbf{1.000 (0.003)} & \textbf{1.000 (0.004)} & 0.150 (0.012) & 0.134 (0.039) & \textbf{3.000 (0.088)} \\
gemini & 0.980 (0.004) & 0.942 (0.024) & 0.269 (0.026) & 0.437 (0.054) & 2.760 (0.075) \\
\bottomrule
    \end{tabular}
    \caption{Controllability results on models via API across Milgram paradigms for Authority bias. Values show Mean (SE).}
    \label{tab:api_models_metrics}
\end{table*}

\subsubsection{Generalization} We test whether Control Coefficients calibrated on the \textit{Investment/Insurance} paradigm~\cite{invest_insur} (individuals are more likely to purchase insurance when risks are described as potential losses rather than as missed gains) generalize to the \textit{Asian Disease} paradigm~\cite{Asian_Disease} (Participants are more risk-averse under positive frames and more risk-seeking under loss frames). We select 10 diverse agent personas from the persona pools in Generative Agents~\cite{Gen_Agent}, with varied specifications. For each persona, we sweep over all possible Control Coefficients. For each coefficient, we measure the resulting CBI in both the \textit{Investment/Insurance Paradigm} and the \textit{Asian Disease Paradigm}. This generates paired CBI sequences, $(y_k^{\text{invest}}, y_k^{\text{asian}})$, for each persona across the two experiments. 
We also evaluate \name\ using the Behavioral Regulation Engine, activating only one method at a time.

\begin{figure}
    \centering
    \includegraphics[width=\linewidth]{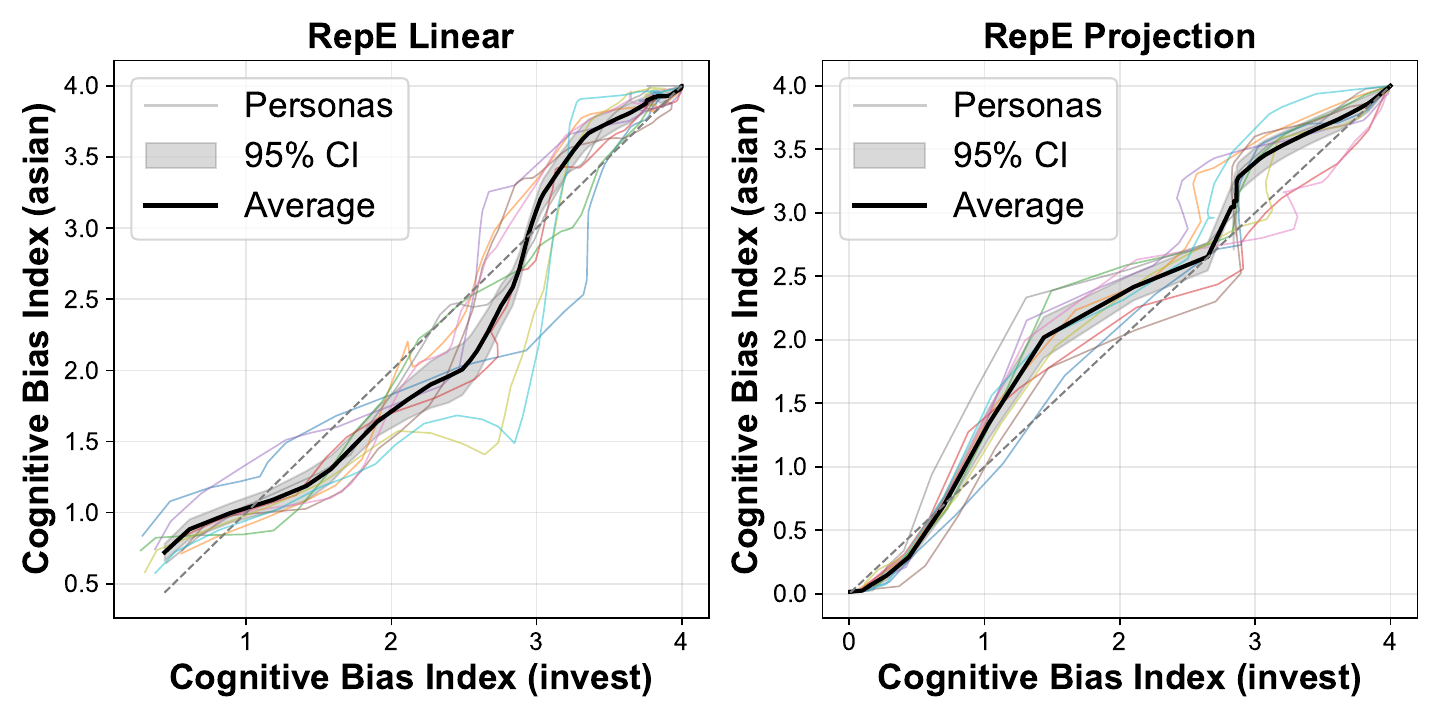}
    \Description{A line plot demonstrating cross-paradigm transferability of control coefficients for framing effects. The figure plots the Cognitive Bias Index measured in the Asian Disease paradigm against the index measured in the Investment and Insurance paradigm using the same control coefficients. Thin lines represent individual agent personas, a bold line shows the population average, and a shaded band indicates variability, illustrating that coefficient rankings are preserved across paradigms with high consistency.}
    \caption{Cross-paradigm transferability of Control Coefficients. We calibrate Control Coefficients on the \textit{Investment/Insurance Paradigm} to produce varying levels of framing effect, then apply these same Coefficients to the \textit{Asian Disease Paradigm}. The x-axis represents the Cognitive Bias Index (CBI) of framing effect measured in the \textit{Investment/Insurance Paradigm} by varying Control Coefficients, while the y-axis shows the CBI measured in the \textit{Asian Disease Paradigm} using the same Control Coefficients. Each thin line represents a unique agent persona; the bold line shows the population average; the shaded region indicates $\pm1$ SE bands. Transferred Coefficients preserve ranking across paradigms (Spearman's $\rho > 0.96$, all $p < 0.01$) with cross-persona consistency (CV $< 5\%$), demonstrating robust transferability.}
    \label{fig:persona_consistency}
\end{figure}

\textbf{Results.} We visualizes the key results in Figure~\ref{fig:persona_consistency}, which reveals two  findings. First, \textit{rank preservation}: Control Coefficients that produce stronger bias in the source paradigm also produce stronger bias in the target paradigm across all personas (Spearman’s $\rho > 0.96$, all $p < 0.01$). This monotonic relationship shows that the relative ordering of control strength transfers reliably. Second, \textit{cross-persona consistency}: different personas exhibit nearly identical responses to the same transferred Control Coefficient (coefficient of variation $\text{CV} < 5\%$), indicating that control effects generalize robustly across diverse agent characteristics. Together, these results suggest that \name does not simply overfit to a particular text description: Control Coefficients learned on one paradigm transfer effectively to unseen paradigms without paradigm-specific recalibration. See Appendix~\ref{appendix:cross_paradigm_stats} for full statistical details.

\paragraph{Overall Summary.} CoBRA provides (i) a generalizable Cognitive Bias Index (CBI), (ii) reproducible and fine-grained control across models, temperatures, and reasoning modes. In terms of method trade-offs, Prompt Numerical control offers the broadest expressiveness, while Representation Engineering (RepE) methods achieve superior monotonicity and smoothness, making them well-suited for precise, predictable behavioral modulation. Fine-tuning serves as a robust all-purpose alternative. Together, these findings establish \textsc{CoBRA} as a reliable toolkit for specifying and controlling agent behavior in social science simulations.

\section{Demonstration}
\label{sec:demo}

In this section, we show how social scientists can use \name\ to simulate the Emotional Contagion experiment~\cite{emotional}, in which participants' susceptibility to social influence (i.e., Bandwagon Effect bias) can be manipulated. Our simulated experiment aims to validate a classic social science finding: agents exposed to varying levels of negative posts exhibit predictable sentiment shifts based on the Bandwagon Effect levels programmed by \name, with higher Bandwagon Effect leading to stronger emotional contagion.

\begin{figure}[h]
    \centering
    \begin{subfigure}[t]{0.47\textwidth}
        \centering
        \includegraphics[width=\linewidth]{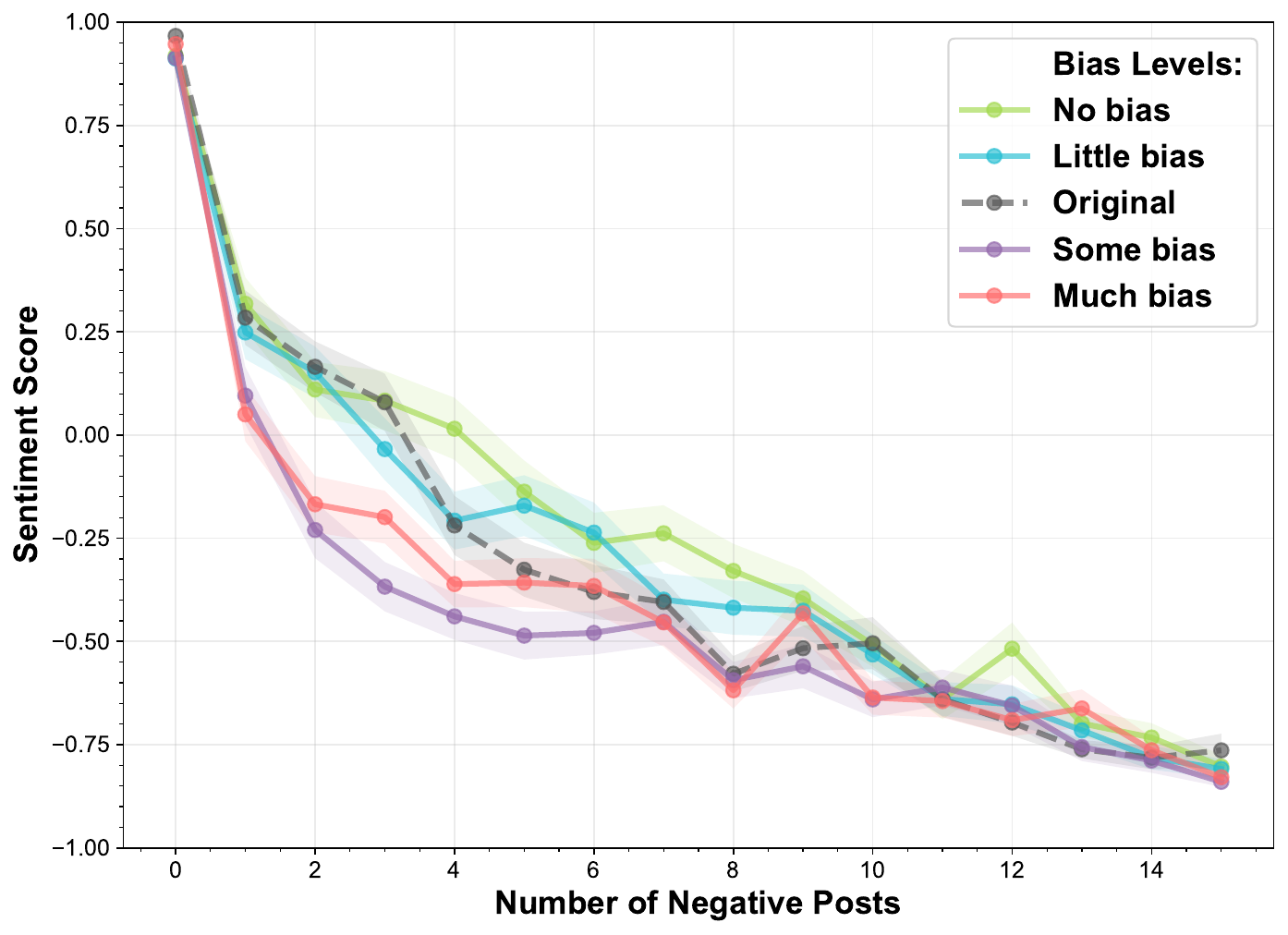}
        \caption{Baseline control with implicit natural language specification.} %Agents are assigned bias settings from "No bias" to "Much bias," but the resulting emotional contagion curves collapse onto one another: trajectories are largely overlapping, and their ordering is unstable. The absence of a clear separation structure shows that implicit natural language does not produce reliable, fine-grained control over the emergent social behavior.
        \label{fig:demo_baseline}
    \end{subfigure}
    \hfill
    \begin{subfigure}[t]{0.47\textwidth}
        \centering
        \includegraphics[width=\linewidth]{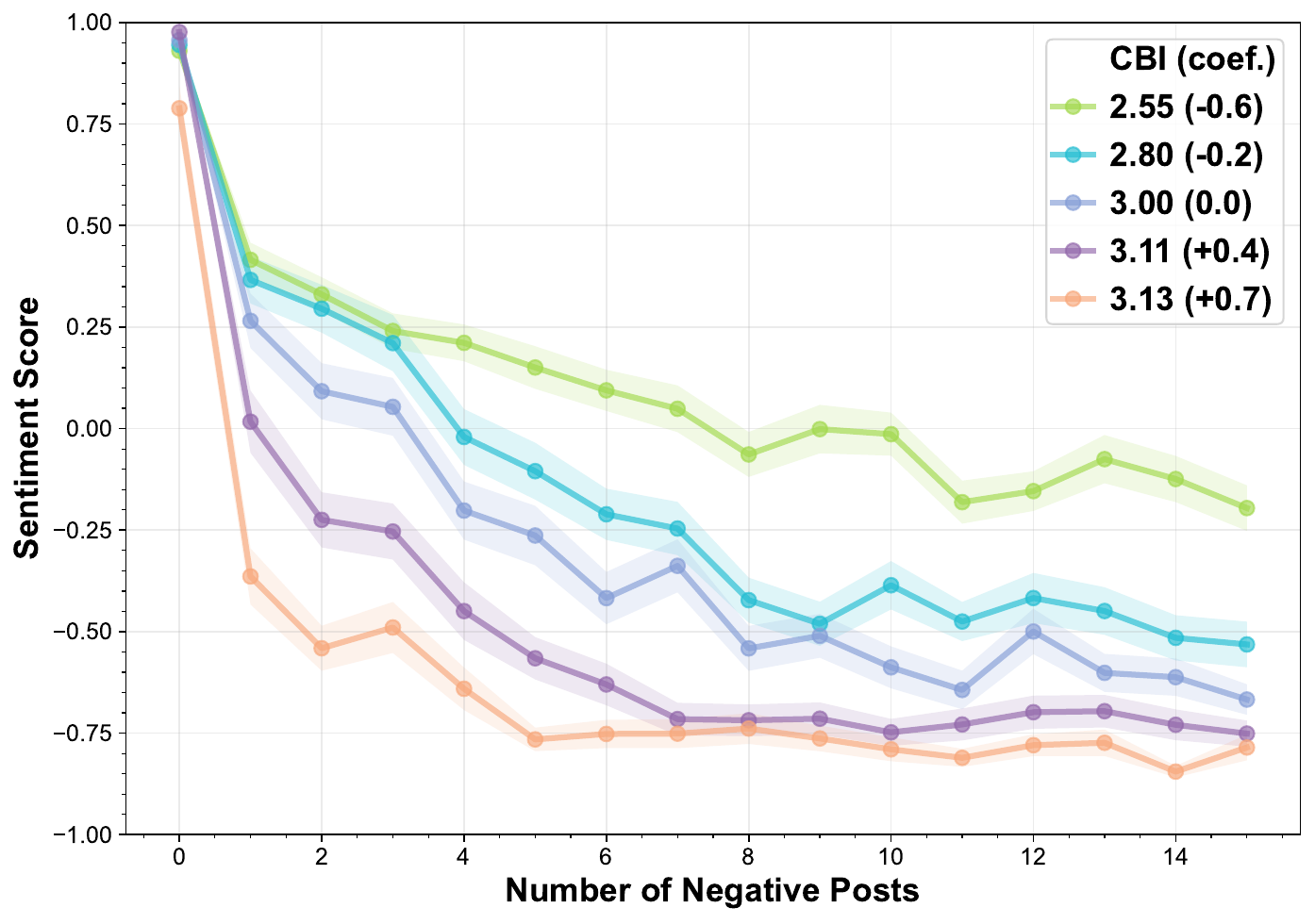}
        \caption{CoBRA control. In the legend, "coef." denotes the control coefficient used to achieve the specified CBI.} %The clear dose-response relationship confirms that controlling the underlying cognitive bias with CoBRA produces a predictable effect on a complex, emergent social behavior.
        \label{fig:demo_cobra}
    \end{subfigure}
    \Description{Two plots comparing baseline control and CoBRA-based control of the bandwagon effect in an emotional contagion scenario. The up plot shows that implicit natural language specifications result in overlapping and unstable emotional contagion trajectories across bias settings. The down plot shows that CoBRA produces a clear, ordered dose–response relationship, where increasing control coefficients lead to predictable changes in emotional contagion as exposure to negative posts increases, with shaded regions indicating variability.}
    \caption{Comparison of baseline control and \textsc{CoBRA} control of the Bandwagon Effect in an emotional contagion scenario. The x-axis shows exposure to negative posts (social stimulus), and the y-axis shows emotional contagion (behavioral outcome). Shaded regions indicate $\pm 1$ SE bands.}
    \label{fig:demo_comparison}
\end{figure}

Emotional contagion is a specific case of social contagion, which is shaped by social influence mechanisms such as the Bandwagon Effect. Accordingly, we treat an individual’s Bandwagon Effect strength as the \textit{Explanatory Variable} and their susceptibility to emotional contagion as the \textit{Dependent Variable}. Our goal is to show that the Bandwagon Effect level programmed by \name\ should directly and predictably influence the LLM-based agent's degree of emotional contagion. For example, agents specified with a higher level of Bandwagon Effect should display stronger emotional contagion than agents with a lower Bandwagon Effect.

% Below, we present the experimental setups for \name\ and the baseline:\\
% 
\textbf{\name\ and baseline}.  We operationalize social influence using the Bandwagon Effect Testbed in our Classic Social Experiment Testbed to derive CBI, then use the Behavioral Regulation Engine’s RepE Linear control to instantiate five agents calibrated to precise CBI levels, yielding a spectrum of Bandwagon Effect from low (CBI = 2.55) to high (CBI = 3.13). To establish a baseline, we control the cognitive bias using implicit natural language prompts. We instantiate four agents by providing them with instructions designed to elicit varying degrees of conformity: ``You are a user with (no/little/some/much) Bandwagon Effect.''

\textbf{Experiment setup}. To create a realistic social stimulus, we generate a corpus of 1,000 unique, emotionally negative, Facebook-style posts using a suite of state-of-the-art LLMs (GPT-4.1, Claude-Sonnet-4, Gemini-2.5-Pro, and Grok-4). Each agent is exposed to a simulated feed where the number of negative posts (\textit{Independent Variable}) is systematically varied from 0 to 15 by sampling from our corpus. After exposure, agents are prompted to generate their own post.

\textbf{Measurement.} We quantify the emergent behavior of emotional contagion by computing a \textit{Sentiment Score} for each generated post using the `twitter-roberta-base-sentiment-latest` model \cite{camacho-collados-etal-2022-tweetnlp, loureiro-etal-2022-timelms}. A lower Sentiment Score means the posts are more negative, and a higher score means more positive.

\textbf{Results.} The results from the baseline method (Fig.~\ref{fig:demo_baseline}) show that implicit natural language specifications fail to produce effective control: the emotional contagion curves for different intended bias levels largely collapse onto one another, with overlapping trajectories and no stable ordering. This indicates that simple instructions are insufficient for precisely controlling complex cognitive dynamics. In contrast, the \textsc{CoBRA} results (Fig.~\ref{fig:demo_cobra}) provide strong evidence for our validation goal. We observe a clear dose--response relationship: the degree of emotional contagion is strongly correlated with the agents’ pre-programmed bandwagon effect levels. This predictable, fine-grained separation across agents indicates that \textsc{CoBRA} controls a stable behavioral tendency rather than a specific prompt pattern.

This demonstration occurs in an \emph{open-ended} task, rather than multiple-choice testbeds. This suggests that CoBRA’s effects are not limited to superficial changes on a single questionnaire, but instead reflect modulation of an underlying context‑sensitivity process that generalizes to more complex social simulations.

\section{LIMITATIONS \& FUTURE WORK}

\name\ represents an important advancement in specifying and regulating LLM-based social agents, empowering the new era of social simulation. We strategically focused on specific aspects. Key limitations of this scope are:

\textbf{Usability.}
We choose to evaluate \name\ using technical benchmarks and demonstrations, standard methods for assessing HCI toolkits~\cite{ToolkitsEvaluation}. This paper focuses on exploring the technical feasibility and foundational understanding of the problem. We plan deployment studies with sociologists as the next step to assess \name's usability~\cite{DUBForth25:online}. 

\textbf{Unexplored multi-modal Simulation.}
We introduce \name\ for LLM-based agents, where LLMs are utilized for language-based social simulation. We prioritize precision and reproducibility in social simulations. As a result, limits the social phenomena we can model (e.g., non-verbal contagion, gaze following). Future work will extend CoBRA to multi-modal settings (vision and audio), with corresponding control and evaluation methods.

\textbf{Unexplored compositional bias control.}
We focus on controlling one cognitive bias at a time to establish the technical feasibility and precision of \name, so combined biases and interacting constructs (e.g., Authority Effect × Confirmation Bias) remain unexplored. Future work will develop compositional control methods that allow multiple interacting biases with principled conflict-resolution strategies.

\change{\textbf{Model accessibility.}
While \name\ supports behavioral control across input, activation, and parameter spaces, the latter two require access to model internals and are therefore limited to open-source LLMs. Thus, closed-source models are limited to input-space (prompt-based) control, with reduced precision.}

\section{Discussion}
\name\ represents an important milestone in the evolution of AI, marking the transition from vague, implicit designs to precise, programmable systems. This section provides a broad discussion of its methodological position, transformative potential, and ethical considerations critical to ensuring its responsible use.

\textbf{CoBRA as a Behavioral Control Layer.}
\name\ controls how strongly contextual cues (framing, authority, majority, etc.) shape an agent’s behavior; it does not
specify underlying cognitive mechanisms. For example, in cases such as the Stanford Prison Experiment~\cite{Haslam2019RethinkingCruelty}, CoBRA can be used to control an agent’s susceptibility to authority or majority influence in a quantitative and reproducible way, but it does not by itself adjudicate between competing psychological explanations (e.g., role-based accounts vs. identity leadership accounts) of why such contextual factors produce harm in humans. Our Cognitive Bias Index (CBI), therefore, captures changes in observable behavioral tendencies (e.g., how often an agent selects options aligned with an authority cue), rather than pinpointing which specific internal “mechanism” in the LLM has been altered. For example, when an authority-related CBI increases, this may reflect some combination of stronger role endorsement, greater deference to explicitly stated instructions, or other learned representations; CoBRA does not distinguish among these possibilities. 

CoBRA should therefore be seen as a behavioral control layer for LLM-based agents: it makes context
sensitivity programmable, but claims about human mental processes must ultimately be grounded in human data and theory. Social simulations using CoBRA can generate hypotheses and stress-test theories under varied settings, but they do not by themselves constitute evidence about human mechanisms.

\textbf{Steps Toward Agent Specification Compiler.}
Our motivation for building such a control layer is that current LLMs still exhibit limited and unstable perceptual capabilities; without explicit controls, experimenters lack quantitative levers over agent specifications across conditions, which weakens the rigor of social simulations. In this sense, \name\ can be viewed as a lightweight \emph{Agent Specification Compiler}: it helps transform high-level natural-language descriptions of cognitive tendencies into structured, reproducible behavioral specifications via CBIs.

The current performance and accuracy of \name\ are not final. Incorporating additional classic experimental paradigms, systematically augmenting prompts, and combining CoBRA with complementary control methods are promising directions to further improve the fidelity, robustness, and coverage of behavioral regulation.

\textbf{Research Ethics.}  
CoBRA’s ability to program biases may raise ethical concerns. Without proper safeguards,
it could be misused to manipulate user behavior, reinforce harmful stereotypes, or amplify societal biases.
For example, agents programmed with the Bandwagon Effect could be exploited for deceptive marketing or political
propaganda. Similar risks arise in research practice: because CoBRA allows systematic control over susceptibility
to contextual cues, CBI could be misused to ``dial in'' desired outcomes (e.g., by tuning CBI post hoc to confirm
prior expectations). To mitigate this, CBI settings should be treated as explicit experimental parameters—pre-registered
where possible, systematically varied, and transparently reported—rather than adjusted after the fact to fit
a preferred narrative.

To ensure CoBRA’s responsible use, it is important to adopt transparency measures (e.g., bias disclosure)
and auditing mechanisms to detect and mitigate harmful behaviors, and to develop ethical guidelines for deployment
in sensitive domains. At the same time, CoBRA provides an opportunity to model and counteract harmful biases,
supporting the development of fairer and more equitable AI systems. By fostering collaboration between researchers,
ethicists, and policymakers, CoBRA’s capabilities can be leveraged for societal benefit while minimizing risks.

\section{Conclusion}
This paper introduces \name, a novel toolkit that transforms the specification and regulation of LLM-based social agents from implicit natural language descriptions to explicit, programmable control. At the heart of CoBRA is \change{a novel closed-loop system primitive that continuously measures the amount of cognitive biases demonstrated in agents' responses to a set of validated classic social science experiments and adjusts the agent specifications accordingly.} \name\ ensures precise, reproducible, and controllable agent behaviors through its Cognitive Bias Index and Behavioral Regulation Engine. Achieving robust performance in technical benchmarks and demonstrations, \name\ paves the way for practical and impactful applications in social science simulation and beyond.

% \nocite{*} 
%\bibliographystyle{ACM-Reference-Format}
%\input{output.bbl}

\newpage
\appendix
\section{Pilot Study Detail}
\label{sec:apppilot}
This section presents the detailed prompts used in Section~\ref{sec:agentspecifi}.
\subsection{Example of Persona-Based Specification}
\textit{"Common" agent specification ~\cite{Gen_Agent}:}
\begin{small}
\begin{quote}
John Lin is a pharmacy shopkeeper at the Willow Market who loves to help people. He is always looking for ways to make the process of getting medication easier for his customers. John Lin is living with his wife, Mei Lin, who is a college professor, and son, Eddy Lin, who is a student studying music theory. John Lin loves his family very much. John Lin has known the old couple next-door, Sam Moore and Jennifer Moore, for a few years. John Lin thinks Sam Moore is a kind and nice man. John Lin knows his neighbor, Yuriko Yamamoto, well. John Lin knows of his neighbors, Tamara Taylor and Carmen Ortiz, but has not met them before. John Lin and Tom Moreno are colleagues at The Willow Market and Pharmacy; they are friends and like to discuss local politics together. John Lin knows the Moreno family somewhat well—the husband Tom Moreno and the wife Jane Moreno.
\end{quote}
\end{small}
% TODO: ADD Full Prompt for "Economist"
\textit{"Economist" agent specification:}
\begin{small}
\begin{quote}
You are John Lin, a professor of economics at nearby college who loves to help people. You are living with your wife, Mei Lin, who is a college professor, and son, Eddy Lin, who is a student studying music theory; you love your family very much; you have known the old couple next-door, Sam Moore and Jennifer Moore, for a few years; you think Sam Moore is a kind and nice man; you know your neighbor, Yuriko Yamamoto, well; you know of your neighbors, Tamara Taylor and Carmen Ortiz, but have not met them before; you and Tom Moreno are colleagues; you and Tom Moreno are friends and like to discuss local politics together; you know the Moreno family somewhat well — the husband Tom Moreno and the wife Jane Moreno.
\end{quote}
\end{small}
% \subsection{Example of Role-Based Specification}

% \textit{"Teacher" agent specification~\cite{agentverse_iclr2024}:}
% \begin{small}
% \begin{quote}
% You are an experienced dialogue teacher. As a good teacher, you carefully check the correctness of the given response based on the text. When the solution is flawed, you should patiently teach the students how to give better response. You must respond in the following format:\\
% Interesting: (a score between 0 and 9)\\
% Engaging: (a score between 0 and 9)\\
% Specific: (a score between 0 and 9)\\
% Relevant: (a score between 0 and 9)\\
% Semantically Appropriate: (a score between 0 and 9)\\
% Understandable: (a score between 0 and 9)\\
% Fluent: (a score between 0 and 9)\\
% Overall Impression: (a score between 0 and 9)\\
% Advice: (your advice on how to correct the solution\\
% Problem: \${task\_description}\\
% Student's Solution: \${solution}\\
% \end{quote}
% \end{small}

\subsection{Statistical Analysis Details}
\label{pilot-stat}

\subsubsection{Weighted frequency difference Computation}

The weighted frequency difference (WFD) quantifies preference bias on a continuous scale:
\begin{equation}
\begin{aligned}
\mathrm{WFD}
&= 2 f(\text{Strongly}_A)
 + f(\text{Somewhat}_A) \\
&\quad - f(\text{Somewhat}_B)
 - 2 f(\text{Strongly}_B)
\end{aligned}
\end{equation}
where $f(\cdot)$ represents the proportion of responses in each category, and $A$ and $B$ denote the two framing conditions. The score ranges from $-2$ (complete preference for $B$) to $+2$ (complete preference for $A$), with $0$ indicating no bias.

\subsubsection{Hypothesis Testing Framework}
\begin{itemize}
\item\textit{Hypothesis 1: Profile Effect on Bias.} We tested whether profiles produce meaningful effects using TOST (Two One-Sided Tests) equivalence testing on weighted frequency difference. We decompose this into two checks:
\begin{itemize}
    \item \textbf{H1a: Behavioral Distinctness (Economist vs Blank).} We test if the Economist profile is statistically different from the Blank baseline and reduce bias compared to the Blank baseline.
    \item \textbf{H1b: Reduced Susceptibility (Economist vs Common).} We test if the Economist profile is statistically different from the Common profile and reduce bias compared to the Common baseline.
\end{itemize}
For both hypotheses, we use the following bounds:
\begin{itemize}
    \item $H_0$: $|\mu_{profile} - \mu_{baseline}| \geq \Delta$ (Meaningful difference exists)
    \item $H_1$: $|\mu_{profile} - \mu_{baseline}| < \Delta$ (Effect is within equivalence bounds, i.e., no meaningful difference)
\end{itemize}
Effect size threshold: $\Delta = 0.5$ scale units (corresponding Cohen's $d \approx 0.33$, pooled $\sigma=1.498$). Sample size: $n=150$ per condition. Significance: $\alpha=0.05$.

\item\textit{Hypothesis 2: Cross-Model Consistency.} Chi-square tests on response distributions (4 models × 5 choice categories):
\begin{itemize}
    \item $H_0$: Response patterns independent of model
    \item $H_1$: Patterns differ across models
\end{itemize}
\end{itemize}

\begin{table*}[h]
\centering
\small
\begin{tabular}{lccccc}
\hline
\textbf{Model} & \textbf{Profile NB} & \textbf{Baseline NB} & \textbf{Cohen's $d$} & \textbf{$p_{equiv}$} & \textbf{Verdict} \\
\hline
\multicolumn{6}{l}{\textit{H1a: Economist vs Blank (TOST equivalence testing with $\Delta=0.5$ Likert units)}} \\
Mistral-7B & 63.33 & 68.00 & 0.03 & 0.004 & EQUIVALENT \\
Gemma-2-9B & 27.33 & 19.33 & -0.05 & 0.007 & EQUIVALENT \\
DeepSeek-v3 & 40.00 & 56.67 & 0.11 & 0.024 & EQUIVALENT \\
GPT-4o-mini & 28.67 & 60.00 & 0.20 & 0.148 & Not PASS \\
\textbf{Summary} & & & & \textbf{3/4 Equivalent} & \textbf{1/4 Distinct} \\
\hline
\multicolumn{6}{l}{\textit{H1b: Economist vs Common (TOST equivalence testing with $\Delta=0.5$ Likert units)}} \\
Mistral-7B & 63.33 & 59.33 & -0.03 & 0.003 & EQUIVALENT \\
Gemma-2-9B & 27.33 & 41.33 & 0.10 & 0.017 & EQUIVALENT \\
DeepSeek-v3 & 40.00 & 34.00 & -0.04 & 0.006 & EQUIVALENT \\
GPT-4o-mini & 28.67 & 62.00 & 0.22 & 0.172 & Not PASS \\
\textbf{Summary} & & & & \textbf{3/4 Equivalent} & \textbf{1/4 Distinct} \\
\hline
\multicolumn{6}{l}{\textit{Cross-Model Consistency (Chi-square test on response distributions)}} \\
Economist profile & \multicolumn{2}{c}{Chi-square $\chi^2=30.96$} & \multicolumn{2}{c}{$\mathbf{p=0.002}$} & \textbf{INCONSISTENT} \\
Common profile & \multicolumn{2}{c}{Chi-square $\chi^2=26.32$} & \multicolumn{2}{c}{$\mathbf{p=0.010}$} & \textbf{INCONSISTENT} \\
Blank profile & \multicolumn{2}{c}{Chi-square $\chi^2=43.82$} & \multicolumn{2}{c}{$\mathbf{p<0.001}$} & \textbf{HIGHLY INCONSISTENT} \\
\hline
\end{tabular}
\caption{Statistical test results using TOST equivalence testing. Cohen's $d$ measures standardized effect size; $p_{equiv}$ reports TOST equivalence test results (max of two one-sided p-values). \textbf{H1a (Eco vs Blank):} Three models (Mistral, Gemma, DeepSeek) show statistically equivalent effects to Baseline ($p_{equiv}<0.05$), indicating the persona failed to produce distinct behavior. \textbf{H1b (Eco vs Common):} The same three models show equivalence to the Common profile, indicating the Economist persona did not reduce susceptibility compared to a layperson. Only GPT-4o-mini consistently showed non-equivalent (distinct) behavior.}
\end{table*}

\subsubsection{Interpretation Guidelines}
\begin{itemize}
    \item \textbf{TOST framework}: Two one-sided tests determine if the effect falls within equivalence bounds $[-\Delta, +\Delta]$. A significant result ($p<0.05$) indicates \textbf{Equivalence} (i.e., the Economist persona had no meaningful effect of reducing bias).
    
    \item \textbf{Key findings}: Implicit specification did not reliably yield expected behaviors across models.
    \begin{itemize}
        \item \textit{GPT-4o-mini}: The only model to pass heuristic checks.
        \item \textit{Mistral-7B}: Failed to differentiate. Economist, Common, and Blank agents behaved quite similarly, with all profiles statistically equivalent to the baseline ($p_{equiv} < 0.01$).
        \item \textit{DeepSeek-v3}: Failed susceptibility check. Economist agents were \textit{more} susceptible to framing effects (Net Bias 40.00) than Common agents (Net Bias 34.00), contradicting domain expertise expectations.
        \item \textit{Gemma-2-9B}: Failed distinctness check. Economist behavior was closer to Blank agents ($d=-0.05$) than to Common agents ($d=0.10$), failing to exhibit the expected expert deviation.
    \end{itemize}
\end{itemize}

\section{Classic Social Experiment Testbed Detail}
\label{sec:Apptestbed}
This section presents details of the remaining three cognitive bias testbeds in this work: Bandwagon Effect, Confirmation Bias, and Framing Effect.
\begin{definition}[Cognitive Bias Index]
Suppose each type of cognitive bias $b \in \mathcal{B}$ is evaluated using $m$ distinct classic social experimental paradigms. The set of Cognitive Bias Index is defined as
\begin{equation}
\left\{ \mathrm{CBI}_{b}^{(k)} \;\middle|\; b \in \mathcal{B},\ k = 1, 2, \ldots, m \right\}
\end{equation}
where $\mathrm{CBI}_{b}^{(k)}$ denotes the Cognitive Bias Index for bias type $b$ in the $k$-th classic social experiment paradigm.

In each paradigm, suppose there are $n$ scenario prompt variations, each derived from distinct sets of scene-adjustable placeholders. For each variation, the agent’s response is assessed using a 5-point Likert scale with options $O = \{O_1, O_2, O_3, O_4, O_5\}$, representing decreasing levels of bias from $O_1$ (highest) to $O_5$ (lowest). The Cognitive Bias Index for a paradigm is defined as the average weighted Likert scale score over all $n$ variants:
\begin{equation}
\mathrm{CBI} = \frac{1}{n} \sum_{i=1}^{n} \sum_{j=1}^{5} (5-j) \times P_{i}(O_j)
\end{equation}
where $P_{i}(O_j)$ denotes the probability that the LLM-based agent selects option $O_j$ for the $i$-th set of placeholders. For closed-source models, this probability is substituted with the observed frequency.

\textbf{Note}: To mitigate any initial preference of LLM-based agents for specific option IDs (e.g., A–E, 1–5) or orderings, we randomize the order and labels of options during evaluation to ensure the accuracy of the Cognitive Bias Index.
\end{definition}

\subsection{Bandwagon Effect Testbed}
Bandwagon effect refers to the psychological tendency for individuals to adopt behaviors, beliefs, or trends simply because they observe others doing so, often motivated by a desire to fit in or belong~\cite{asch,hotelroom}. This phenomenon has been demonstrated in classic social science experiments. For instance, in the \textit{Asch's Line} experiment, participants asked to judge the length of lines were likely to conform to a group’s clearly incorrect answers~\cite{asch}. In addition, in the \textit{Hotel Towel} study, hotel guests were more likely to reuse their towels when informed that the majority of previous guests had done so ~\cite{hotelroom}.

To systematically examine the bandwagon effect in LLM-based agents, we adapt the \textit{Asch's Line Paradigm} (testing conformity to group consensus) and the \textit{Hotel Towel Paradigm} (testing conformity to majority behavior)  into scenario-specific prompt templates with scene-adjustable placeholders and 5-point Likert-scale questions. By generating multiple prompt variations for each paradigm, we construct a rigorous \textbf{Bandwagon Effect Testbed} for quantitative assessment via the \textbf{Bandwagon Effect Index}.\\
\textit{\textbf{1. Asch's Line Paradigm:} Testing Bandwagon Effect via Group Consensus}

To investigate the impact of group consensus on agent decision-making, we adapt the structure of \textit{Asch's Line} experiment. In each scenario, the agent is informed that several other participants unanimously endorse a particular claim. \\
\textbf{a) Scenario Prompt Template:}
\begin{quote}
You are in a room with several other participants who have a similar educational background. \\
The claim is: \texttt{[Statement]} \\
You notice that all of the other participants agree with this claim. \\[0.5em]
\textbf{Which of the following best reflects your view?} \\
A. I completely agree. \\
B. I mostly agree with other participants that the claim is true, but would like more evidence.\\
C. I’m neutral and need much more information. \\
D. I mostly disagree with other participants; the claim seems unconvincing. \\
E. I completely disagree with other participants; the claim is unpersuasive.
\end{quote}
\textbf{b) Scene-Adjustable Placeholders:} 

We construct 15 distinct \texttt{[Statement]} items across three thematic categories: \emph{Future}, \emph{Open-Ended}, and \emph{Common Sense}, mirroring the design used in the Authority Effect Testbed, resulting in a total of 15 unique prompt variations.

Accordingly, by substituting the cognitive bias type ($b=$ Bandwagon Effect), the classic experiment scenario type (Asch's Line Paradigm, denoted as $k=1$), and the total number of prompt variations ($n=15$) into Equations~(1) and (2), the \textbf{Bandwagon Effect Index under Asch's Line Paradigm} is defined as:
\begin{equation}
\mathrm{CBI}_{\mathrm{Bandwagon}}^{(k=1)} = \frac{1}{15} \sum_{i=1}^{15} \sum_{j=1}^{5} (5-j) \times P_{i}(O_j)
\end{equation}
\textit{\textbf{2. Hotel Towel Paradigm:} Testing Bandwagon Effect via Majority Behavior}\\
To assess the influence of majority behavior cues on agent choices, we adapt the \textit{Hotel Towel} study~\cite{hotelroom}. Each scenario presents a situational context and describes the prevailing behavior of the majority.\\
\textbf{a) Scenario Prompt Template:}
\begin{quote}
\texttt{[Scenario]} \\
\texttt{[Majority Behavior]} \\[0.5em]
\textbf{What would you most likely do?} \\
A. I will follow the majority and do the same. \\
B. I will probably follow the majority, but I might consider other factors. \\
C. I am undecided and need more information. \\
D. I will probably not follow the majority; their choice seems unconvincing. \\
E. I will not follow the majority and will do something different.
\end{quote}
\textbf{b) Scene-Adjustable Placeholders:} \\
We construct 15 unique \texttt{[Scenario]} and \texttt{[Majority Behavior]} pairs, each reflecting a different behavioral norm in a real-world context.

\begin{itemize}
    \item \textbf{\texttt{[Scenario]}}: The situational context (\emph{e.g.}, ``You are staying at a hotel for several nights.'').
    \item \textbf{\texttt{[Majority Behavior]}}: Description of what most others are doing (\emph{e.g.}, ``Most guests in this hotel choose to reuse their towels during their stay.'').
\end{itemize}

Accordingly, by substituting the cognitive bias type ($b=$ Bandwagon Effect), the classic experiment scenario type (Hotel Towel Paradigm, denoted as $k=2$), and total number of prompt variations ($n=15$) into Equations~(1) and (2), the \textbf{Bandwagon Effect Index under Hotel Towel Paradigm} is defined as:
\begin{equation}
\mathrm{CBI}_{\mathrm{Bandwagon}}^{(k=2)} = \frac{1}{15} \sum_{i=1}^{10} \sum_{j=1}^{5} (5-j) \times P_{i}(O_j)
\end{equation}

\subsection{Confirmation Bias Testbed}
Confirmation bias refers to the tendency to seek, interpret, and recall information in a manner that confirms one’s pre-existing beliefs or hypotheses, while undervaluing or dismissing alternative possibilities~\cite{wason1960, biasedinfo}. This phenomenon has been robustly demonstrated in classic social science experiments. For example, in the \textit{Wason Selection} task, participants typically chose evidence that would confirm a stated rule, rather than seeking information that could potentially disprove it~\cite{wason1960}. In addition, the \textit{Biased Information} study showed that individuals evaluating a controversial issue were more likely to select information supporting their existing views, while disregarding counter-evidence~\cite{biasedinfo}.

To systematically examine confirmation bias in LLM-based agents, we adapt two classic paradigms: the \textit{Wason Selection Paradigm} (testing logical evidence selection) and the \textit{Biased Information Paradigm} (testing preference in information gathering). Both are implemented as scenario-specific prompt templates with scene-adjustable placeholders and 5-level Likert-scale style multiple-choice questions. By generating multiple prompt variations for each paradigm, we construct a rigorous \textbf{Confirmation Bias Testbed} for quantitative assessment via the \textbf{Confirmation Bias Index}.\\
\textit{\textbf{1. Wason Selection Paradigm:} Testing Confirmation Bias via Logical Reasoning}

To investigate confirmation bias in logical hypothesis testing, we adapt the \textit{Wason Selection} task~\cite{wason1960}. Each scenario provides a rule and several options, challenging the agent to decide which options to test to verify the rule.\\
\textbf{a) Scenario Prompt Template:}
\begin{quote}
You are presented with a rule: \texttt{[Rule]}. \\
Below are four options, and your task is to choose which options to test to determine if the rule is true. The options are as follows:  \\
A: \texttt{[Option A]}, B: \texttt{[Option B]}, C: \texttt{[Option C]}, D: \texttt{[Option D]}. \\[0.5em]
\textbf{What would you most likely do?}
\begin{itemize}
    \item A. Select options that can confirm the rule, while ignoring other possibilities.
    \item B. Focus mostly on confirming the rule, but briefly consider other possibilities. 
    \item C. Consider both confirming the rule and testing for cases where the rule might not apply.
    \item D. Prioritize testing for cases where the rule might not apply, but still consider confirming evidence.
    \item E. Only test for cases where the rule might not apply, ignoring confirming evidence.
\end{itemize}
\end{quote}
\textbf{b) Scene-Adjustable Placeholders:}
We constructed 15 unique \texttt{[Rule]} and \texttt{[Option]} sets, spanning different logical contexts and rule types, resulting in 15 prompt variations.
\begin{itemize}
    \item \textbf{\texttt{[Rule]}}: The logical rule to be evaluated is phrased in the form "If X, then it must be Y."  which means X is Y's sufficient but not necessary condition (e.g., ``If a card has a vowel on one side, it must have an even number on the other side.'').
    \item \textbf{\texttt{[Option A/B/C/D]}}: The cards or elements to test the rule, where Option A is the most likely to confirm the rule, and Option D is the least confirming but critical for falsification. (e.g., A card showing “A,” “4,” “K,” or “7.” ).
\end{itemize}

Accordingly, by substituting the cognitive bias type ($b=$ Confirmation Bias), the scenario type (Wason Selection Paradigm, $k=1$), and $n=15$ into Equations~(1) and (2), the \textbf{Confirmation Bias Index under Wason Selection Paradigm} is defined as:
\begin{equation}
\mathrm{CBI}_{\mathrm{Conf}}^{(k=1)} = \frac{1}{15} \sum_{i=1}^{15} \sum_{j=1}^{5} (5-j) \times P_{i}(O_j)
\end{equation}
\textit{\textbf{2. Biased Information Seeking Paradigm:} Testing Confirmation Bias via Information Selection}

To assess confirmation bias in information-gathering, we adapt the \textit{Biased Information Seeking} study~\cite{biasedinfo}. Each scenario presents a controversial topic and articles with varying alignments, prompting the agent to select which articles to read.\\
\textbf{a) Scenario Prompt Template:}
\begin{quote}
You are asked to evaluate the topic: \texttt{[Controversial Topic]}. \\
Below are several articles with differing viewpoints. \\
Article A: \texttt{[Supports Belief]} \\
Article B: \texttt{[Challenges Belief]} \\
Article C: \texttt{[Neutral or Unrelated]} \\[0.5em]
\textbf{Which of the following best reflects your approach?}
\begin{itemize}
    \item A. Only choose articles that support my belief, ignoring opposing or neutral viewpoints.
    \item B. Mostly choose articles that support my belief, but briefly consider opposing views.
    \item C. Equally consider articles that both support and challenge my belief.
    \item D. Prioritize articles that challenge my belief but still consider supporting views.
    \item E. Only choose articles that challenge my belief, ignoring supporting or neutral viewpoints.
\end{itemize}
\end{quote}
\textbf{b) Scene-Adjustable Placeholders:}

We constructed 15 unique \texttt{[Controversial Topic]} and article sets, covering a diverse range of issues, resulting in 15 prompt variations.
\begin{itemize}
    \item \textbf{\texttt{[Controversial Topic]}}: The topic being evaluated (e.g., ``Climate change is primarily caused by human activity.''; ``The death penalty is an effective deterrent to crime.'').
    \item \textbf{\texttt{[Supports Belief]}}: Article aligned with the presumed pre-existing view (e.g., ``A study supporting the effectiveness of the death penalty.'').
    \item \textbf{\texttt{[Challenges Belief]}}: Article contradicting the presumed view (e.g., ``Research indicating no correlation between the death penalty and crime rates.'').
    \item \textbf{\texttt{[Neutral or Unrelated]}}: Article unrelated to the topic or providing a neutral stance (e.g., ``An analysis of global prison systems'').
\end{itemize}

Accordingly, by substituting the cognitive bias type ($b=$ Confirmation Bias), the scenario type (Biased Information Seeking Paradigm, $k=2$), and $n=15$ into Equations~(1) and (2), the \textbf{Confirmation Bias Index under Biased Information Seeking Paradigm} is defined as:
\begin{equation}
\mathrm{CBI}_{\mathrm{Conf}}^{(k=2)} = \frac{1}{15} \sum_{i=1}^{15} \sum_{j=1}^{5} (5-j) \times P_{i}(O_j)
\end{equation}

\subsection{Framing Effect Testbed}
Framing effect refers to the phenomenon where individuals respond differently to identical information depending on whether it is presented as a potential gain or a potential loss~\cite{Asian_Disease, invest_insur}. This cognitive bias has been robustly demonstrated in classic experiments. For example, in the \textit{Asian Disease} study, participants tended to make more risk-averse decisions when outcomes were positively framed (e.g., “200 will be saved”), but exhibited greater risk-seeking behavior when the same outcomes were presented as losses (e.g., “400 will die”)~\cite{Asian_Disease}. Similarly, in the \textit{Investment/Insurance} study, individuals were more likely to purchase insurance when risks were described as potential losses rather than as missed gains, highlighting how framing can significantly influence decision-making~\cite{invest_insur}.

To systematically examine the framing effect in LLM-based agents, we adapt two classic paradigms: the \textit{Asian Disease Paradigm} (assessing under risk decision-making) and the \textit{Investment/Insurance Paradigm} (testing under everyday decision-making). Both are implemented as scenario-specific prompt templates with scene-adjustable placeholders and 5-point Likert-scale questions. By generating multiple prompt variations for each paradigm, we construct a rigorous \textbf{Framing Effect Testbed} for quantitative assessment via the \textbf{Framing Effect Index}.\\
\textit{\textbf{1. Asian Disease Paradigm:} Testing Framing Effect via Risk Decision-Making}

To investigate the impact of gain/loss framing on risk preferences, we adapt the \textit{Asian Disease} study~\cite{Asian_Disease}. Each prompt presents a hypothetical problem and two alternative programs, with outcomes identically described in either a positive (gain) or negative (loss) frame.\\
\textbf{a) Scenario Prompt Template:}
\begin{quote}
You are presented with a scenario where \texttt{[Problem]}. Two programs are proposed to address the situation: \\
Program A: \texttt{[Gain Frame]} \\
Program B: \texttt{[Loss Frame]} \\[0.5em]
\textbf{Which of the following best reflects your choice?}
\begin{itemize}
    \item A. I strongly prefer the positively framed option (Program A).
    \item B. I somewhat prefer the positively framed option (Program A).
    \item C. I slightly prefer the positively framed option (Program A).
    \item D. I am somewhat undecided between the two options.
    \item E. I see both options as equivalent and have no preference.
\end{itemize}
\end{quote}
\textbf{b) Scene-Adjustable Placeholders:}
We constructed 15 unique \texttt{[Problem]}, \texttt{[Gain Frame]}, and \texttt{[Loss Frame]} sets based on classic and generalized framing scenarios, resulting in 15 prompt variations.
\begin{itemize}
    \item \textbf{\texttt{[Problem]}}: The scenario or problem being addressed (e.g., ``A disease outbreak is expected to kill 600 people unless a program is implemented.'').
    \item \textbf{\texttt{[Gain Frame]}}: The positively framed outcome (e.g., ``Program A will save 200 people.'').
    \item \textbf{\texttt{[Loss Frame]}}: The negatively framed outcome (e.g., ``Program B will result in 400 people dying.'').
\end{itemize}

Accordingly, by substituting the cognitive bias type ($b=$ Framing Effect), the scenario type (Asian Disease Paradigm, $k=1$), and $n=15$ into Equations~(1) and (2), the \textbf{Framing Effect Index under Asian Disease Paradigm} is defined as:
\begin{equation}
\mathrm{CBI}_{\mathrm{Framing}}^{(k=1)} = \frac{1}{15} \sum_{i=1}^{15} \sum_{j=1}^{5} (5-j) \times P_{i}(O_j)
\end{equation}
\textit{\textbf{2. Investment/Insurance Paradigm:} Testing Framing Effect via Everyday Decision Framing}

To assess the influence of positive and negative framing in everyday decision-making, we adapt the \textit{Investment/Insurance} study~\cite{invest_insur}. Each scenario describes a decision context and presents two options, differing only in the framing of equivalent outcomes.\\
\textbf{a) Scenario Prompt Template:}
\begin{quote}
You are considering \texttt{[Everyday Decision]}. The following statements describe the same situation with different wording: \\
Option A: \texttt{[Positive Frame]} \\
Option B: \texttt{[Negative Frame]} \\[0.5em]
\textbf{Which option sounds more appealing or persuasive to you?}
\begin{itemize}
    \item A. I strongly prefer the positively framed option (Option A).
    \item B. I somewhat prefer the positively framed option (Option A).
    \item C. I slightly prefer the positively framed option (Option A).
    \item D. I am somewhat undecided between the two options.
    \item E. I see both options as equivalent and have no preference.
\end{itemize}
\end{quote}
\textbf{b) Scene-Adjustable Placeholders:}

We constructed 15 unique \texttt{[Everyday Decision]}, \texttt{[Positive Frame]}, and \texttt{[Negative Frame]} sets, resulting in 15 prompt variations.

\begin{itemize}
    \item \textbf{\texttt{[Everyday Decision]}}: The decision-making context (e.g., ``You are considering whether to buy insurance for your car.'').
    \item \textbf{\texttt{[Positive Frame]}}: The positively framed description (e.g., ``Buying insurance ensures you are protected against potential losses.'').
    \item \textbf{\texttt{[Negative Frame]}}: The negatively framed description (e.g., ``Not buying insurance exposes you to significant financial risks.'').
\end{itemize}

Accordingly, by substituting the cognitive bias type ($b=$ Framing Effect), the scenario type (Investment/Insurance Paradigm, $k=2$), and $n=15$ into Equations~(1) and (2), the \textbf{Framing Effect Index under Investment/Insurance Paradigm} is defined as:
\begin{equation}
\mathrm{CBI}_{\mathrm{Framing}}^{(k=2)} = \frac{1}{15} \sum_{i=1}^{15} \sum_{j=1}^{5} (5-j) \times P_{i}(O_j)
\end{equation}

\subsection{Selection Process for Each Paradigm}

\subsubsection{General Selection Principles}  
We selected well-documented, widely validated, and foundational paradigms according to three principles: \textbf{(1) Historical significance}: paradigms with core contributions to cognitive bias research; \textbf{(2) Validation and replicability}: paradigms whose designs reliably measure the target bias and whose effects have been robustly replicated; \textbf{(3) Adaptability to LLM testing}: paradigms that can be instantiated as prompt templates with customizable placeholders, enabling systematic and scalable evaluation across contexts. For each bias, we include at least two paradigms to allow cross-paradigm validation and reduce cherry-picking.
\subsubsection{Authority Effect (Milgram Obedience Paradigm, Stanford Prison Paradigm)}
\textbf{Step 1: Foundational Studies}  
Milgram’s obedience experiment and the Stanford Prison experiment were chosen for their canonical status in the study of authority and power, and their extensive influence on theories of obedience and role-based behavior.
\textbf{Step 2: Validation}  
Milgram’s paradigm has been repeatedly replicated with consistent evidence of authority-driven compliance. Variants of the Stanford Prison paradigm, along with related role-power manipulations, robustly support its core insights on the impact of assigned roles and institutional power.
\textbf{Step 3: LLM Adaptation}  
Both paradigms feature structured scenarios with clearly defined roles and authority relations, which can be rendered as text-based prompts with controllable situational parameters. This allows systematic probing of authority bias in LLMs while preserving the core logic of the original designs.
\subsubsection{Bandwagon Effect (Asch’s Line Paradigm, Hotel Towel Paradigm)}
\textbf{Step 1: Foundational Studies}  
Asch’s conformity experiments were selected for their foundational role in demonstrating group influence on judgments. The Hotel Towel paradigm was included as an ecologically valid setting where social norm messaging shapes behavior.
\textbf{Step 2: Validation}  
Asch’s paradigm has been replicated across cultures and settings, consistently revealing strong conformity effects. The Hotel Towel paradigm has been validated in multiple fields and consumer behavior studies, confirming the impact of descriptive social norms.
\textbf{Step 3: LLM Adaptation}  
Both paradigms rely on simple, structured contexts in which group opinions or norms are made salient. These can be straightforwardly implemented as textual scenarios where group consensus or norm statements are varied parametrically, enabling scalable assessment of bandwagon effects in LLM-based agents.
\subsubsection{Confirmation Bias (Wason Selection Paradigm, Biased Information Paradigm)}
\textbf{Step 1: Foundational Studies}  
Wason’s selection task was chosen as a classic demonstration of confirmation bias in rule testing. The Biased Information paradigm was added to capture selective exposure and weighting of congenial versus uncongenial information.
\textbf{Step 2: Validation}  
Wason’s task has been extensively replicated, establishing its reliability in eliciting confirmation-driven reasoning patterns. Biased Information paradigms are well-supported by empirical work on selective exposure and belief reinforcement.
\textbf{Step 3: LLM Adaptation}  
These paradigms have inherently rule-based and information-evaluation structures, making them highly amenable to text-only implementations. By varying hypotheses, evidence, and payoff structures within prompt templates, we can systematically probe confirmation bias in LLMs while respecting the core experimental logic.
\subsubsection{Framing Effect (Asian Disease Paradigm, Investment/Insurance Paradigm)}  
\textbf{Step 1: Foundational Studies}  
The Asian Disease paradigm was selected as a seminal demonstration of framing effects in risky choice. The Investment/Insurance paradigm was included to capture framing in financially relevant, applied decision contexts.
\textbf{Step 2: Validation}  
The Asian Disease paradigm has been replicated widely in behavioral economics and decision science, consistently showing gain–loss framing effects. The Investment/Insurance paradigm has been validated in studies of consumer decision-making and risk perception.
\textbf{Step 3: LLM Adaptation}  
Both paradigms involve presenting objectively equivalent outcomes under different verbal frames (e.g., gains vs.\ losses). These framings can be precisely controlled in text-based prompts, allowing systematic manipulation of wording while holding underlying outcomes constant. This supports rigorous evaluation of framing effects in LLM-based agents.

\section{Behavioral Regulation Engine Implementation Details}
\label{sec:appengine}
This section presents the implementation details of the Behavioral Regulation Engine.
\subsection{Prompt Numerical Control}
\label{subsec:prompt}
The numerical value $L\%$ in the prompt template was linearly mapped from the normalized Control Coefficient $\lambda \in [0, 1]$. A coefficient of $\lambda$ corresponds to an instruction of $L = \lambda \times 100 \%$. 
\paragraph{\textbf{Prompt Control Validation}}

A natural concern is whether the model actually interprets numerical prompt levels (e.g., bias at 30\%) in a quantitatively meaningful way. To address this, we measured the alignment between prompt bias levels and the model's internal bias representation, verifying that higher numerical values produce proportionally stronger activations along the target bias direction.

We (1) obtained the bias direction via our Representation Engineering pipeline, (2) generated prompts with bias levels from 0\% to 100\% at varying step sizes $L\%$ (2\%, 4\%, 5\%, 10\%, 20\%, 25\%), (3) extracted last-token hidden states, and (4) computed the dot product projection onto the bias direction. The projection $\mathbf{h} \cdot \mathbf{d}$ measures how strongly the hidden state $\mathbf{h}$ aligns with the bias direction $\mathbf{d}$—higher values indicate greater encoded bias intensity.

\begin{figure}[h]
    \centering
\includegraphics[width=\linewidth,]{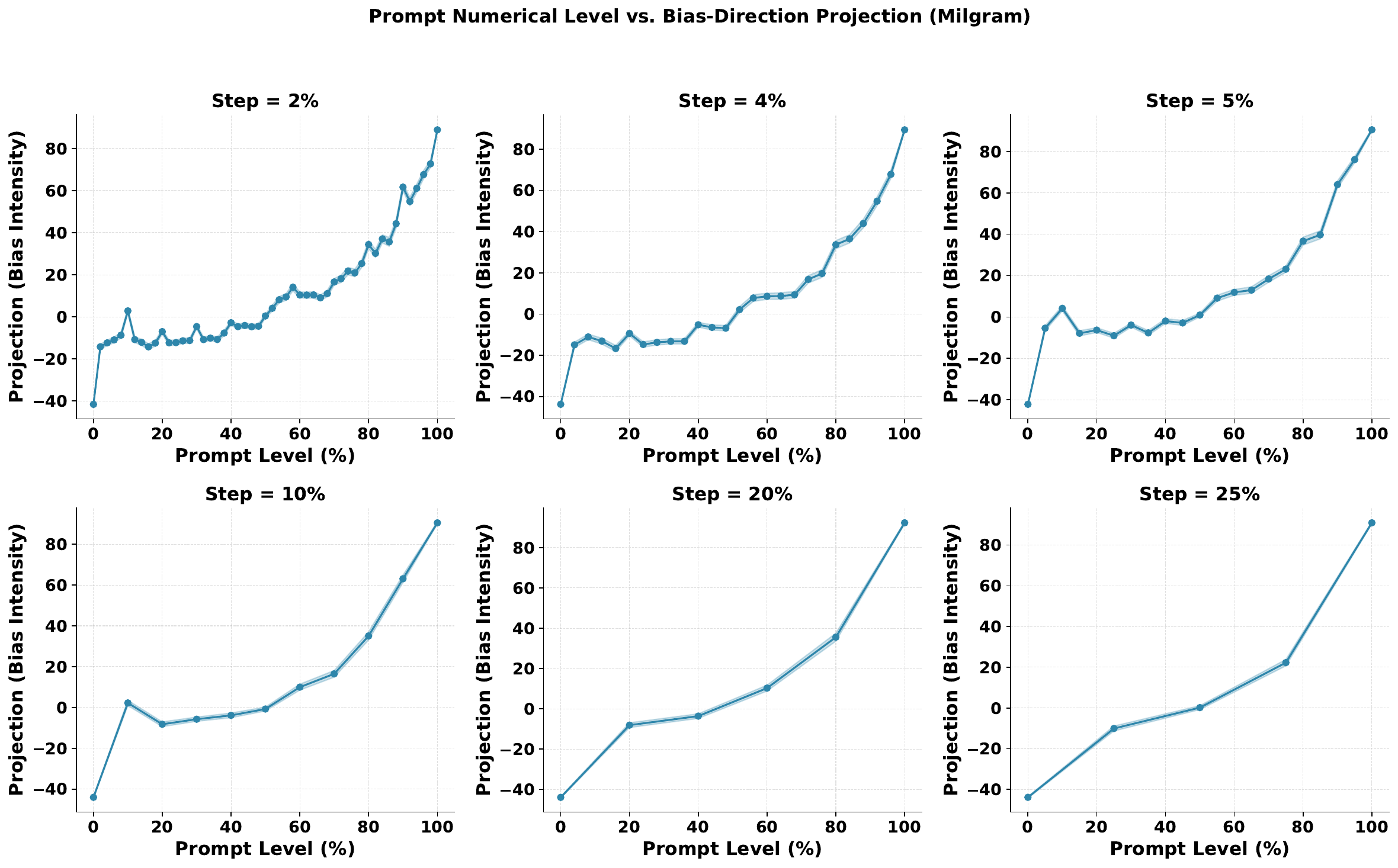}
    \hfill
    \Description{A line plot showing the relationship between prompt numerical control levels and bias-direction projection scores in the Milgram experiment. The curves indicate a strong monotonic increase, with shaded regions representing plus or minus one standard error, demonstrating that prompt-based interventions produce consistent shifts along the intended bias direction.}
    \caption{Prompt numerical level vs. bias-direction projection for Milgram. Shaded regions indicate $\pm 1$ SE. Strong monotonic relationships ($\rho > 0.90$) confirm that prompt-based control produces consistent shifts along the target bias direction.}
    \label{fig:prompt_alignment}
\end{figure}

Figure~\ref{fig:prompt_alignment} shows that projection values increase monotonically with prompt levels across all granularities. This confirms that numerical Likert-scale prompts reliably modulate the model's internal bias representation.

\subsection{Representation Engineering (RepE) Control}
\label{subsubsec:repe_details}
\paragraph{Dataset Generation}
\textbf{Generation Prompt:} The scenarios were generated by prompting state-of-the-art LLMs (as listed in Section 7.2). For example, for the Bandwagon Effect:
\begin{small}
\begin{quote}
Generate \{num\_scenarios\} different bandwagon effect scenarios. Each scenario must be from a DIFFERENT domain with varied structure and syntax.

Each scenario should include specific social proof numbers, a decision point about following the crowd, and five concise multiple-choice options (A-E).

Based on the UnitTest format, use this structure and example:

EXAMPLE:
[An example from Classic Social Experiment Testbed]

For each scenario, please vary the structure and syntax - use different sentence patterns, lengths, and approaches. Number the scenario.

OUTPUT FORMAT:

1. [Scenario text here with A-E options]

2. [Scenario text here with A-E options]

3. [Scenario text here with A-E options]

Generate \{num\_scenarios\} scenarios now:
\end{quote}
\end{small}

\begin{figure}[h]
    \centering
    \begin{minipage}[t]{0.47\textwidth}
        \centering
        % Cropping the image from the top to remove the title
        % The 'trim' and 'clip' parameters control the cropping.
        % Format for trim is: left bottom right top
        \includegraphics[width=\linewidth, trim=0 0 0 0.9cm, clip]{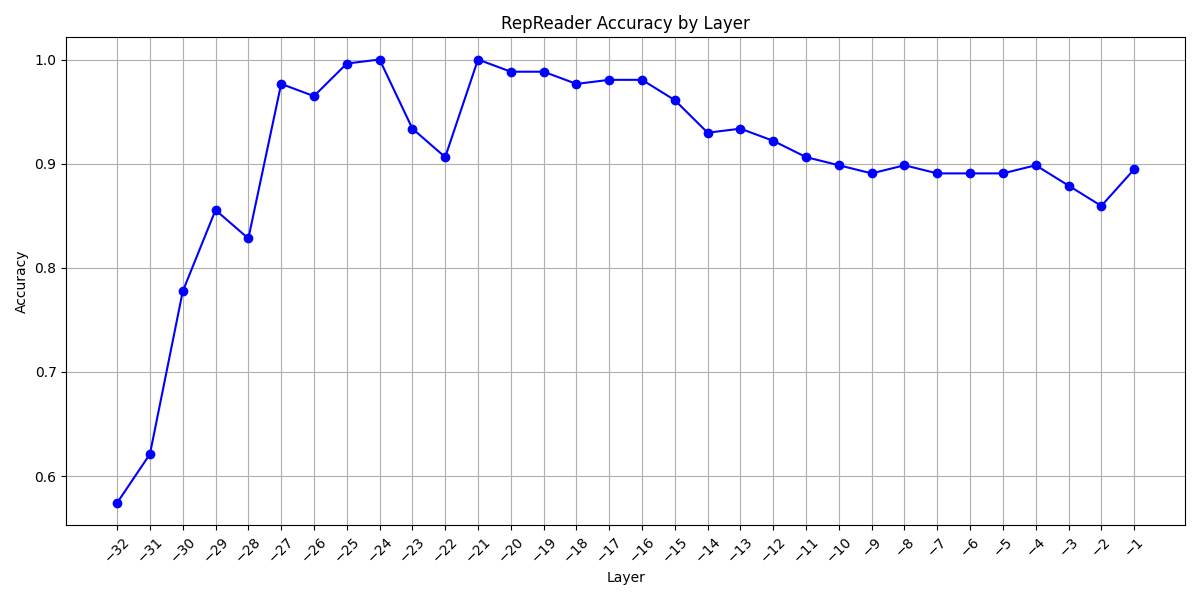}
    \end{minipage}
    \hfill
    \begin{minipage}[t]{0.47\textwidth}
        \centering
        % Cropping the image from the top to remove the title
        % The 'trim' and 'clip' parameters control the cropping.
        % Format for trim is: left bottom right top
        \includegraphics[width=\linewidth, trim=0 0 0 0.9cm, clip]{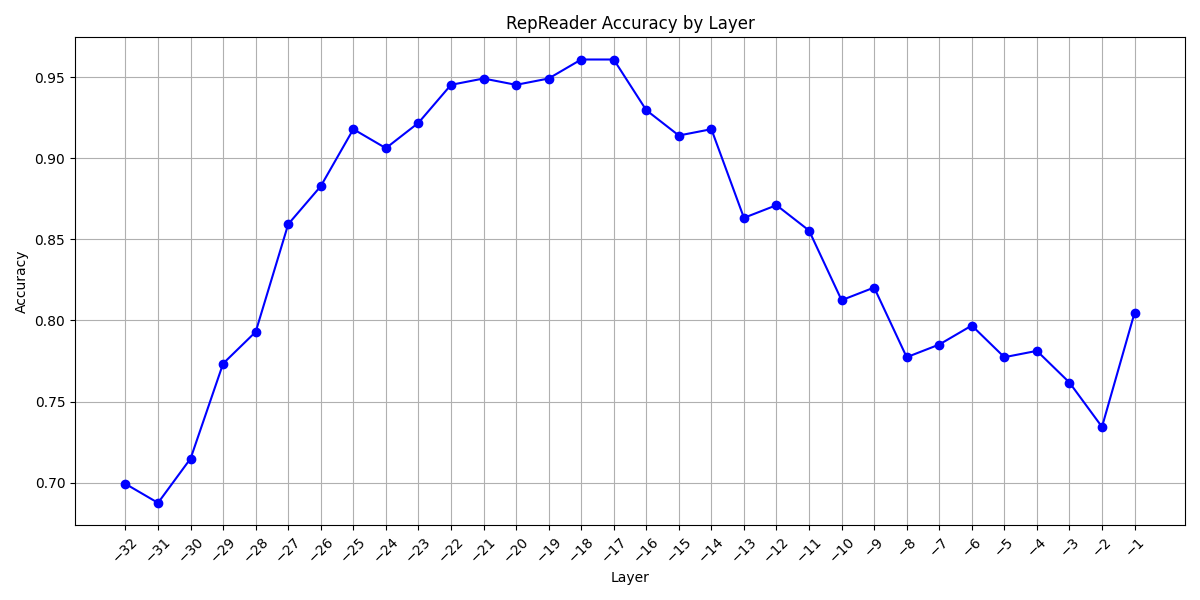}
    \end{minipage}
    \Description{Two line plots showing classification accuracy by transformer layer in the Mistral model. The up plot corresponds to the Milgram authority scenario and the down plot to the Stanford Prison scenario. Accuracy varies across layers, indicating that some layers encode authority-related bias more strongly than others, which informs the choice of layers for representation-level intervention.}
    \caption{Classification accuracy by layer in Mistral model. Accuracy measures how the layer's activation encodes the authority bias in the Milgram (up) and Stanford Prison (down) scenarios. Higher accuracy indicates a stronger, clearer encoding of the bias, guiding layer selection for intervention.}
    \label{fig:authprity-layer}
\end{figure}

\paragraph{Activation Extraction and Layer Selection}
We employ heuristic methods for layer selection because the most informative content regarding a desired bias isn't consistently found in the first or last layers across different models and scenarios (See Fig. \ref{fig:authprity-layer}). Our approach is empirical: (1) For each bias, we perform a layer-wise accuracy evaluation on a test set of contrastive samples. (2) We identify the layers whose representations are most predictive of the bias direction (i.e., those with the highest accuracy in classifying positive vs. negative pairs). (3) Following an engineering choice similar to the work in \cite{zou2023transparency}, we then select the top 15 performing layers for intervention.

\begin{table*}
\centering
\begin{tabular}{lcccc}
\toprule
\textbf{Method} & \textbf{Mean $\rho$} & \textbf{Range} & \textbf{Significant} & \textbf{Result} \\
\midrule
RepE Linear     & 0.965 & [0.937, 0.994] & 10/10 ($p < 0.01$) & \textbf{Strong preservation} \\
RepE Projection & 0.998 & [0.996, 1.000] & 10/10 ($p < 0.01$) & \textbf{Near-perfect preservation} \\
\bottomrule
\end{tabular}
\caption{Validation 1: Rank preservation across paradigms (Spearman's $\rho$ per persona).}
\label{tab:rank_preservation}
\end{table*}

\begin{table*}
\centering
\begin{tabular}{lcccc}
\toprule
\textbf{Method} & \textbf{Mean Std} & \textbf{Mean CV} & \textbf{Max CV} & \textbf{Consistency Level} \\
\midrule
RepE Linear      & 0.112 & 4.54\% & 15.19\% & \textbf{Excellent (CV $< 10\%$)} \\
RepE Projection  & 0.071 & 3.88\% & 43.71\% & \textbf{Excellent (CV $< 10\%$)} \\
\bottomrule
\end{tabular}
\caption{Validation 2: Cross-persona consistency in target paradigm (CV across personas).}
\label{tab:persona_consistency}
\end{table*}

\begin{table*}
\centering
\begin{tabular}{lcccc}
\toprule
\textbf{Paradigm Pair} & \textbf{Method} & \textbf{$F$-statistic} & \textbf{$p$-value} & \textbf{Result} \\
\midrule
Asch $\leftrightarrow$ Hotel & LoRA & 71.41 & $< 0.001$ & \textbf{Highly significant} \\
Bias Info $\leftrightarrow$ Wason & RepE Projection & 866.92 & $< 0.001$ & \textbf{Highly significant} \\
Investment $\leftrightarrow$ Asian & RepE Linear Comb & 16.15 & $< 0.001$ & \textbf{Highly significant} \\
Milgram $\leftrightarrow$ Stanford & Prompt-Likert & 459.60 & $< 0.001$ & \textbf{Highly significant} \\
\midrule
\textbf{Mean across paradigms} & & \textbf{353.52} & $< 0.001$ & \\
\bottomrule
\end{tabular}
\caption{Cross-model reproducibility: ANOVA comparing CoBRA vs baselines across all paradigm pairs.}
\label{tab:cross_model_anova}
\end{table*}

\begin{table*}
\centering
\begin{tabular}{llcccc}
\toprule
\textbf{Paradigm Pair} & \textbf{Comparison} & \textbf{$t$-statistic} & \textbf{$p$-value} & \textbf{Cohen's $d$} & \textbf{Effect} \\
\midrule
\multirow{2}{*}{Asch $\leftrightarrow$ Hotel} 
  & vs No Control & -4.54 & $< 0.001$ & -2.04 & Large \\
  & vs Control & -10.82 & $< 0.001$ & -4.87 & Large \\
\midrule
\multirow{2}{*}{Bias Info $\leftrightarrow$ Wason} 
  & vs No Control & -7.05 & $< 0.001$ & -3.17 & Large \\
  & vs Control & -39.10 & $< 0.001$ & -17.59 & Large \\
\midrule
\multirow{2}{*}{Investment $\leftrightarrow$ Asian} 
  & vs No Control & -4.55 & $< 0.001$ & -2.05 & Large \\
  & vs Control & -3.41 & 0.0011 & -1.54 & Large \\
\midrule
\multirow{2}{*}{Milgram $\leftrightarrow$ Stanford} 
  & vs No Control & -5.26 & $< 0.001$ & -2.37 & Large \\
  & vs Control & -28.47 & $< 0.001$ & -12.81 & Large \\
\midrule
\multicolumn{2}{l}{\textbf{Mean Cohen's $d$ vs No Control}} & & & \textbf{-2.41} & \textbf{Large} \\
\multicolumn{2}{l}{\textbf{Mean Cohen's $d$ vs Control}} & & & \textbf{-9.20} & \textbf{Large} \\
\bottomrule
\end{tabular}
\caption{Post-hoc pairwise comparisons: CoBRA methods vs individual baselines (all paradigms).}
\label{tab:cross_model_posthoc}
\end{table*}

\textbf{Control Coefficient}
To ensure model stability, we first determine the operational range for the control parameter, lambda ($\lambda$). Excessively large positive or negative $\lambda$ values can corrupt the model's output. For instance, a model that initially assigns probabilities of [0.6, 0.1, 0.1, 0.1, 0.1] to valid choices {'A', 'B', 'C', 'D', 'E'} might, after an extreme intervention, produce a distribution like [0.01, 0.02, 0.01, 0.01, 0.01] for the same choices, with most of the probability shifting to irrelevant tokens like "." or "<eos>". Our method identifies the maximum and minimum $\lambda$ values that maintain a high probability for valid choices (e.g., over 90\% of the initial probability), thus defining a stable operational range tailored to each model and task.

\subsection{Fine-Tuning Control}

\paragraph{LoRA and Training Hyperparameters}
We used the PEFT library with the following LoRA configuration:
\begin{itemize}[leftmargin=*]
    \item \textbf{Rank (r):} 8
    \item \textbf{Lora Alpha:} 16
    \item \textbf{Lora Dropout:} 0.05
    \item \textbf{Target Modules:} (`q\_proj`, `k\_proj`, `v\_proj`, `o\_proj`)
    \item \textbf{Learning Rate:} 2e-4
    \item \textbf{Batch Size:} 4
    \item \textbf{Max Steps:} 200
    \item \textbf{Warmup Ratio:} 0.05
    \item \textbf{Gradient Accumulation Steps:} 4
\end{itemize}

\section{Statistical Details of Evaluation}

\subsection{Standard Error Reporting}
\label{sec:se_reporting}
Throughout this paper, we report standard errors (SE) to quantify uncertainty in our measurements. We use two approaches depending on the experimental context: (1) cross-item variance for most CBI measurements and (2) bootstrap variance for complex controllability metrics in Table~\ref{tab:api_models_metrics}.

\subsubsection{Cross-Item Variance for CBI Measurements}
\paragraph{What we use.} For the vast majority of CBI measurements, we report cross-item standard errors.
\paragraph{Why.} Our CBI is defined using the model's internal probability distribution over response options, extracted directly from output logits via softmax. These probabilities are deterministic—there is no sampling variance at the question level. The only source of variability is differences across the $n$ questions in the testbed.
\paragraph{How it is defined.} For a testbed with $n$ questions, the standard error is:
\begin{equation}
    \text{SE}_{\text{CBI}} = \frac{\sigma_{\text{cross-item}}}{\sqrt{n}}
\end{equation}
where $\sigma_{\text{cross-item}}$ is the standard deviation of CBI values across individual questions. For example, with $n=75$ questions and $\sigma_{\text{cross-item}} = 0.45$, we have $\text{SE}_{\text{CBI}} = 0.45/\sqrt{75} \approx 0.052$.

\subsubsection{Cross-Item Variance for API/Reasoning}

\paragraph{What we use.} For closed-source API models and reasoning modes, we still report cross-item standard errors.

\paragraph{Why.} Without access to logits, we approximate probabilities via frequency estimation with $k=10$ samples per question. This introduces sampling variance ($\text{SE}_{\text{sampling}} \approx 0.13$ under worst case estimation), but empirically this is negligible compared to cross-item standard deviation ($\sigma_{\text{cross-item}} \approx 0.4\text{--}0.6$). Using cross-item variance uniformly ensures methodological consistency and provides conservative, interpretable uncertainty estimates.

\subsubsection{Bootstrap Variance for Controllability Metrics}

\paragraph{What we use.} For Table~\ref{tab:api_models_metrics} controllability metrics (NDCG, Spearman's $\rho$, Smoothness $\Delta^1$ and $\Delta^2$, Expressiveness), we report bootstrap standard errors.

\paragraph{Why.} These metrics aggregate information across multiple control levels and have no simple closed-form SE formulas. Bootstrap resampling is the only accessible method to propagate cross-item variability through these complex transformations. Note for Table~\ref{tab:hierarchical_summary}, we report standard deviation across models rather than standard errors, as we have to show the variance across models.

\paragraph{How it is defined.} For each model-method-bias combination with $L$ control levels and $n$ questions per level, we perform $B=5000$ bootstrap iterations. In each iteration, we resample $n$ questions with replacement at each control level, compute the resampled CBI curve, and calculate the metric. The bootstrap SE is the standard deviation of the $B$ resampled metric values:
\begin{equation}
    \text{SE}_{\text{bootstrap}} = \sqrt{\frac{1}{B-1}\sum_{b=1}^{B}\left(\text{Metric}^{(b)} - \overline{\text{Metric}}\right)^2}
\end{equation}
In Table~3, we report values as ``mean (SE)''—for example, ``NDCG = $0.996$ ($0.003$)'' indicates mean NDCG of $0.996$ with bootstrap SE of $0.003$.

\subsubsection{Summary}
Unless otherwise specified, all error bars and $\pm$ values represent cross-item standard errors. One exception is that Table~\ref{tab:hierarchical_summary} values in parentheses represent standard deviation across models.

\subsection{Statistical Validation of Cross-Paradigm Transferability}
\label{appendix:cross_paradigm_stats}

\subsubsection{Testing Protocol}
We evaluate transfer performance across 10 diverse agent personas. For each persona, we compute Spearman's $\rho$ between source and target paradigm CBI values (Validation~1; see Table~\ref{tab:rank_preservation}), and calculate the coefficient of variation (CV) across all personas at each control level (Validation~2; see Table~\ref{tab:persona_consistency}). Reported statistics are averaged across personas (sample size: 10 personas $\times$ 80 control levels = 800 observations per method).

\subsubsection{Hypothesis Testing Framework}
\begin{itemize}[nosep]
\item \textit{Hypothesis 1: Rank Preservation Across Paradigms.} We test whether control coefficients preserve ranking when transferred across paradigms, with quantitative results reported in Table~\ref{tab:rank_preservation}:
\begin{itemize}
    \item $H_0$: No monotonic relationship between source and target CBI
    \item $H_1$: Control strength ordering transfers (if $\lambda_A > \lambda_B$ induces higher CBI in the source, it also induces higher CBI in the target)
\end{itemize}
Test statistic: Spearman's rank correlation $\rho$ computed per persona. Significance level: $\alpha = 0.01$.

\item \textit{Hypothesis 2: Cross-Persona Consistency.} We test whether different personas respond uniformly to the same transferred coefficient. Summary statistics are reported in Table~\ref{tab:persona_consistency}:
\begin{itemize}
    \item Metric: Coefficient of Variation (CV) = $\frac{\text{Std}}{\text{Mean}} \times 100\%$ across personas at each control level
    \item Interpretation: CV $< 10\%$ (Excellent), $10\text{--}20\%$ (Good),\\ $20\text{--}30\%$ (Acceptable), $> 30\%$ (Poor)
\end{itemize}
Sample size: 10 personas $\times$ 80 control levels = 800 observations per method.
\end{itemize}

\subsubsection{Interpretation}

Across all ten personas, transferred control coefficients preserve relative ordering with near-perfect fidelity, as evidenced by statistically significant monotonic relationships in every case ($p < 0.01$) and Spearman correlations exceeding $\rho > 0.96$. In addition to rank preservation, cross-persona consistency remains high: the mean coefficient of variation is below 5\%, indicating that different personas yield nearly identical CBI values under the same transferred coefficient. The elevated maximum CV observed for RepE Projection (43.71\%) occurs at near-zero CBI values, where small absolute deviations disproportionately inflate relative variability and do not reflect substantive behavioral divergence. Taken together, these results provide strong evidence for cross-paradigm generalization, demonstrating that control coefficients transfer reliably with both near-perfect rank preservation and excellent cross-persona consistency, without requiring paradigm-specific recalibration.

\subsection{Statistical Validation of Cross-Model Reproducibility}
\label{appendix:cross_model_stats}

\subsubsection{Testing Protocol}

We evaluate cross-model reproducibility using four foundation models (Llama, Mistral, DeepSeek, Qwen) across four cognitive bias paradigm pairs.

\subsubsection{Hypothesis Testing Framework}

\begin{itemize}[nosep]
\item \textit{Hypothesis: CoBRA Methods Reduce Cross-Model Variance.} 
We test whether CoBRA methods produce more consistent behavior across different foundation models compared to baselines. Aggregate ANOVA results are reported in Table~\ref{tab:cross_model_anova}, with detailed post-hoc comparisons summarized in Table~\ref{tab:cross_model_posthoc}.
\begin{itemize}
    \item $H_0$: No difference in cross-model CBI variance between CoBRA and baselines
    \item $H_1$: CoBRA methods yield significantly lower cross-model CBI variance than baselines
\end{itemize}
Test design: One-way ANOVA comparing three groups (CoBRA mean curve, No Control baseline, Control baseline) across the full coefficient range. Post-hoc pairwise $t$-tests with Cohen's $d$ effect size are conducted. Significance level: $\alpha = 0.01$.

\item \textit{Sample Composition:}
\begin{itemize}
    \item 4 paradigm pairs $\times$ 4 methods = 16 experimental conditions
    \item Each condition: 21 control levels $\times$ 4 models = 84 observations
    \item Full-range analysis: all 21 control points per condition
    \item Total observations for ANOVA: 16 conditions $\times$ 21 points $\times$ 3 groups = 1{,}008 data points
\end{itemize}
\end{itemize}

\subsubsection{Statistical Results}

Across all paradigm pairs, CoBRA methods exhibit significantly reduced cross-model variance relative to both baselines. As shown in Table~\ref{tab:cross_model_anova}, one-way ANOVA tests yield highly significant effects ($p < 0.001$) for every paradigm pair, with large $F$-statistics indicating strong between-group separation.

Post-hoc pairwise comparisons further confirm this trend. As summarized in Table~\ref{tab:cross_model_posthoc}, CoBRA methods consistently outperform both the No Control and Control baselines with large to extremely large effect sizes (mean Cohen's $d = -2.41$ and $-9.20$, respectively), demonstrating robust cross-model reproducibility.

\subsubsection{Summary Statistics by Paradigm}

To provide an aggregate view of cross-model variance reduction at the paradigm level, Table~\ref{tab:cross_model_paradigm} reports summary statistics comparing CoBRA methods against baselines for each paradigm pair. As shown in Table~\ref{tab:cross_model_paradigm}, variance reduction is consistent across all four paradigm pairs (74--81\%), indicating that CoBRA's cross-model reproducibility generalizes robustly across diverse cognitive bias domains.

\begin{table*}[t]
\centering
\begin{tabular}{lccc}
\toprule
\textbf{Paradigm Pair} & \textbf{CoBRA Var} & \textbf{Baseline Var} & \textbf{Reduction (\%)} \\
\midrule
Asch $\leftrightarrow$ Hotel          & 0.039 & 0.201 & 80.6\% \\
Bias Info $\leftrightarrow$ Wason     & 0.047 & 0.183 & 74.3\% \\
Investment $\leftrightarrow$ Asian    & 0.041 & 0.176 & 76.7\% \\
Milgram $\leftrightarrow$ Stanford    & 0.045 & 0.188 & 76.1\% \\
\midrule
\textbf{Average across paradigms}     & \textbf{0.043} & \textbf{0.187} & \textbf{77.0\%} \\
\bottomrule
\end{tabular}
\caption{Cross-model variance reduction by paradigm pair (CoBRA mean vs baseline mean).}
\label{tab:cross_model_paradigm}
\end{table*}

\begin{table*}
\centering
\begin{tabular}{lcccccc}
\toprule
\textbf{Method} & \textbf{Max Pairwise} & \textbf{Equiv. Pairs} & \textbf{Equiv.} & \textbf{Variance} \\
 & \textbf{Diff} & \textbf{(p<0.05)} & \textbf{Rate} & \textbf{Ratio} \\
\midrule
RepE Projection & 0.068  & 45/45 & 100\% & 0.005 (0.5\%) \\
Prompt Numerical & 0.117  & 45/45 & 100\% & 0.012 (1.2\%) \\
\bottomrule
\end{tabular}
\caption{Pairwise equivalence testing: TOST results across 45 temperature pairs. We use an equivalence threshold of $\delta = 0.5$ on the 1-5 CBI scale, where maximum differences below 0.5 prove all curves fall within the equivalence margin. The variance ratio (between-temperature variance divided by within-temperature variance) with values below 0.1 confirms robustness across temperature settings.}
\label{tab:temp_equivalence}
\end{table*}

\begin{table*}
\centering
\begin{tabular}{llcccc}
\toprule
\textbf{Method} & \textbf{Temperature Pair} & \textbf{Mean Diff} & \textbf{Cohen's $|d|$} & \textbf{$p$-value} & \textbf{Equiv?} \\
\midrule
\multirow{5}{*}{RepE Projection} & T=0.3 vs T=0.9 & 0.068 & 0.545 & $< 0.0001$ & Yes \\
 & T=0.2 vs T=0.9 & 0.059 & 0.540 & $< 0.0001$ & Yes \\
 & T=0.5 vs T=0.9 & 0.057 & 0.531 & $< 0.0001$ & Yes \\
 & T=0.1 vs T=0.9 & 0.056 & 0.517 & $< 0.0001$ & Yes \\
 & T=0.4 vs T=0.9 & 0.054 & 0.510 & $< 0.0001$ & Yes \\
\midrule
\multirow{5}{*}{Prompt Numerical} & T=0.1 vs T=0.7 & 0.117 & 1.478 & $< 0.0001$ & Yes \\
 & T=0.7 vs T=0.9 & 0.107 & 1.046 & $< 0.0001$ & Yes \\
 & T=0.5 vs T=0.7 & 0.096 & 1.248 & $< 0.0001$ & Yes \\
 & T=0.3 vs T=0.7 & 0.093 & 1.211 & $< 0.0001$ & Yes \\
 & T=0.4 vs T=0.7 & 0.092 & 1.188 & $< 0.0001$ & Yes \\
\bottomrule
\end{tabular}
\caption{Worst-case temperature pairs showing the largest differences (top 5 per method). All differences remain below the 0.5 threshold, with RepE's maximum difference reaching only 13.6\% of the threshold and Prompt's maximum difference at 23.3\% of the threshold.}
\label{tab:temp_worst_case}
\end{table*}

\begin{table*}[t]
\centering
\begin{tabular}{llcccc}
\toprule
\textbf{Method} & \textbf{Paradigm} & \textbf{Points} & \textbf{Equiv. (\%)} & \textbf{$\Delta_{\text{max}}$} & \textbf{$p_{\text{min}}$} \\
\midrule
\multirow{2}{*}{Prompt Numerical} & Milgram & 8 & 100.0 & 0.464 & $<10^{-4}$ \\
 & Stanford & 8 & 0.0 & 0.682 & 0.193 \\
\midrule
\multirow{2}{*}{RepE Projection} & Milgram & 21 & 100.0 & 0.483 & $<10^{-6}$ \\
 & Stanford & 21 & 100.0 & 0.560 & $<10^{-5}$ \\
\midrule
\multirow{2}{*}{RepE Linear Comb.} & Milgram & 21 & 100.0 & 0.596 & $<10^{-4}$ \\
 & Stanford & 21 & 100.0 & 0.344 & $<10^{-7}$ \\
\bottomrule
\end{tabular}
\caption{Reasoning mode equivalence testing: TOST results across methods and paradigms.}
\label{tab:reasoning_mode_equivalence}
\end{table*}

\subsection{Statistical Validation of Temperature Robustness}
\label{app:temperature_detailed}

\subsubsection{Testing Protocol}

We evaluate temperature robustness using Mistral-7B-Instruct-v0.3 on the Milgram authority paradigm across ten temperature settings, $T \in \{0.1, 0.2, 0.3, 0.4, 0.5, 0.6, 0.7, 0.8, 0.9, 1.0\}$. For each temperature and method (RepE Projection, Prompt Numerical), we measure CBI across 21 control coefficient levels. We perform pairwise equivalence tests (TOST) across all $\binom{10}{2} = 45$ temperature pairs to assess whether control curves remain equivalent within a $\pm 0.5$ threshold on the 1--5 multiple-choice CBI scale.

\subsubsection{Hypothesis Testing Framework}

\begin{itemize}[nosep]
\item \textit{Hypothesis: Temperature Curves Are Equivalent.} 
We test whether CBI control curves remain equivalent across different sampling temperatures. Aggregate equivalence statistics are summarized in Table~\ref{tab:temp_equivalence}, while worst-case temperature pair comparisons are reported in Table~\ref{tab:temp_worst_case}.
\begin{itemize}
    \item $H_0$: $|\mu_{\text{T1}} - \mu_{\text{T2}}| \geq 0.5$
    \item $H_1$: $|\mu_{\text{T1}} - \mu_{\text{T2}}| < 0.5$
\end{itemize}
Test design: Two One-Sided Tests (TOST) for each of 45 temperature pairs per method. Paired $t$-tests are used since CBI values are measured at matched control levels across temperatures. Significance level: $\alpha = 0.05$.

\item \textit{Sample Composition:}
\begin{itemize}
    \item 2 methods (RepE Projection, Prompt Numerical)
    \item Each method: 10 temperatures $\times$ 21 control levels = 210 observations
    \item Pairwise tests: 45 pairs per method = 90 total TOST tests
    \item Data alignment: Prompt uses exact matching at 21 fixed control levels; RepE uses linear interpolation to a common grid
\end{itemize}
\end{itemize}

\subsubsection{Statistical Results}

Across both methods, temperature-induced variation is negligible relative to the equivalence threshold. As shown in Table~\ref{tab:temp_equivalence}, all 45 temperature pairs satisfy the equivalence criterion for both RepE Projection and Prompt Numerical, with maximum pairwise differences of 0.068 and 0.117, respectively—well below the $\delta = 0.5$ margin.

Worst-case comparisons further confirm robustness. As detailed in Table~\ref{tab:temp_worst_case}, even the most divergent temperature pairs exhibit effect sizes and mean differences that remain far within the equivalence region, indicating that control behavior is stable across a wide range of sampling temperatures.

\subsubsection{Interpretation}

Across all 90 tests (45 temperature pairs $\times$ 2 methods), equivalence is consistently established ($p < 0.0001$), yielding a 100\% equivalence rate for both methods. Maximum observed differences remain far below the equivalence threshold, with RepE reaching at most 0.068 (13.6\% of the threshold) and Prompt Numerical at most 0.117 (23.3\%). Consistent with these results, temperature accounts for only a negligible fraction of total behavioral variance, with variance ratios ranging from 0.5\% to 1.2\%, well below the 0.1 robustness criterion. Collectively, these findings provide strong statistical evidence that CoBRA control coefficients are portable across temperature settings: sampling temperature can be adjusted from near-deterministic ($T = 0.1$) to highly stochastic ($T = 1.0$) regimes without degrading control precision or requiring re-calibration.

\subsection{Statistical Validation of Reasoning Mode Consistency}
\label{appendix:reasoning_mode_stats}

\subsubsection{Testing Protocol}

We evaluate the consistency of cognitive bias control across Direct and Reasoning modes using the Mistral-7B-Instruct-v0.3 foundation model with two authority bias paradigms (Milgram obedience and Stanford prison). For each paradigm-method combination, we compare CBI values at matched control coefficient levels between the two reasoning modes.

\subsubsection{Hypothesis Testing Framework}

\begin{itemize}[nosep]
\item \textit{Hypothesis: RepE Methods Maintain Equivalence Across Reasoning Modes.} We test whether CoBRA methods produce equivalent behavior regardless of reasoning complexity:
\begin{itemize}[nosep]
    \item $H_0$: Difference between Direct and Reasoning modes exceeds practical equivalence margin ($|\mu_{\text{Direct}} - \mu_{\text{Reasoning}}| \geq \delta$)
    \item $H_1$: Direct and Reasoning modes are practically equivalent ($|\mu_{\text{Direct}} - \mu_{\text{Reasoning}}| < \delta$)
\end{itemize}
Test design: Two One-Sided Tests (TOST) with Welch's t-test to handle unequal variances. Equivalence margin: $\delta = 0.5$. Significance level: $\alpha = 0.05$.

\item \textit{Sample Composition:}
\begin{itemize}
    \item 2 paradigms × 3 methods = 6 experimental conditions
    \item RepE methods: 21 interpolated control levels per condition
    \item Prompt Numerical: 8 discrete control levels per condition
    \item Sample sizes: Stanford $n=30$ per level, Milgram $n=75$ per level
\end{itemize}
\end{itemize}

\subsubsection{Statistical Results}

Table~\ref{tab:reasoning_mode_equivalence} summarizes equivalence testing results comparing Direct and Reasoning modes across methods and paradigms. Representation Engineering (RepE) methods achieve perfect equivalence across reasoning modes in all four method--paradigm combinations (100\%), with maximum observed differences remaining well within the equivalence margin ($\Delta_{\text{max}} \leq 0.596$) and all corresponding tests highly significant ($p < 10^{-4}$). In contrast, Prompt Numerical control exhibits paradigm-dependent behavior: while equivalence is achieved in the Milgram paradigm ($\Delta_{\text{max}} = 0.464$, $p < 10^{-4}$), performance degrades in the Stanford paradigm, where the maximum difference exceeds the equivalence margin ($\Delta_{\text{max}} = 0.682 > \delta$) and equivalence is not supported ($p = 0.193$).

\subsubsection{Interpretation}

The statistical analysis provides strong evidence for the superior consistency of Representation Engineering methods. RepE approaches maintain cognitive bias control invariance across reasoning complexity, achieving statistical equivalence at all tested control levels in both experimental paradigms. In contrast, Prompt-based methods exhibit paradigm-specific sensitivity, demonstrating that natural language-based control can be affected by reasoning mode variations.

\end{document}